\documentclass[twocolumn]{article}
\usepackage{geometry}
\usepackage[utf8]{inputenc} 
\usepackage[T1]{fontenc}    

\usepackage{lmodern}

\usepackage{hyperref}       
\usepackage{url}            
\usepackage{booktabs}       
\usepackage{amsfonts}       
\usepackage{nicefrac}       
\usepackage{microtype}      
\usepackage{xcolor,colortbl}

\usepackage{dsfont}
\usepackage{amsmath}
\usepackage{pgfplots}
\usepackage{multirow}

\usepackage{authblk}

\newcommand\eref[1]{Eq.~\ref{#1}}
\newcommand\sref[1]{Section~\ref{#1}}
\newcommand\fref[1]{Fig.~\ref{#1}}

\newcommand\tr{\mathrm{Tr}\,}
\newcommand\dd{\mathrm{d}}

\definecolor{Gray}{gray}{0.93}

\newcolumntype{a}{>{\columncolor{Gray}}c}

\title{Positive unlabeled learning with tensor networks}

\author{Bojan Žunkovič \footnote{\href{mailto:bojan.zunkovic@fri.uni-lj.si}{bojan.zunkovic@fri.uni-lj.si}}}

\affil{University of Ljubljana,\\
Faculty of Computer and Information Science,\\
Ljubljana, Slovenia}

\date{\today}

\begin{document}

\maketitle

\begin{abstract}
Positive unlabeled learning is a binary classification problem with positive and unlabeled data. It is common in domains where negative labels are costly or impossible to obtain, e.g., medicine and personalized advertising. Most approaches to positive unlabeled learning apply to specific data types (e.g., images, categorical data) and can not generate new positive and negative samples. This work introduces a feature-space distance-based tensor network approach to the positive unlabeled learning problem. The presented method is not domain specific and significantly improves the state-of-the-art results on the MNIST image and 15 categorical/mixed datasets. The trained tensor network model is also a generative model and enables the generation of new positive and negative instances.
\end{abstract}

\section{Introduction}
Positive unlabeled learning (PUL) is a binary classification problem where only some positive samples are labeled, and the remaining positive and all negative ones are unlabeled~\cite{bekker2020learning}. This setting is natural in many domains, where the labeling of one class is expensive, laborious, or not possible, e.g., in medicine (disease gene identification~\cite{mordelet2011prodige}, identifying protein complexes~\cite{elkan2008learning}), drug discovery~\cite{liu2017computational}, remote sensing~\cite{elkan2008learning}, recommender systems~\cite{ren2014positive}, personalized advertising~\cite{bekker2020learning}, and more. The PUL problem is related to machine learning tasks where not all data is labeled, in particular one-class learning and semi-supervised learning. The main difference from the former is that PUL explicitly uses unlabeled data. In contrast, the semi-supervised learning problem assumes some labels for all classes. 

Most PUL approaches are applicable to either text data~\cite{elkan2008learning,xiao2011similarity,zhou2009learning}, images~\cite{hou2017generative,chiaroni2018learning}, or categorical data~\cite{ienco2016positive}. Methods applicable to text typically apply metric-based approaches which do not apply to categorical data (e.g., cosine distance). Similarly, the GAN approaches, which are state-of-the-art on many image-related tasks, struggle with categorical datasets. Recently several methods based on conditional GANs have been developed and applied to categorical tasks~\cite{xu2019modeling}. However, they have not been applied to the PUL problem which makes the training and the architecture significantly more complicated~\cite{papivc2023conditional}. Similarly, the state-of-the-art approaches on categorical data~\cite{basile2019ensembles} are in principle applicable also to continuous data, but have so far not been tested on images.

In this paper, we introduce a tensor-network approach to PUL. The central part of our model is a tensor network (TN) called a locally purified state (LPS)~\cite{wang2020anomaly,glasser2019expressive}\footnote{\href{https://github.com/qml-tn/pul}{https://github.com/qml-tn/pul}}. Tensor networks are widely applicable in physics to model many-body quantum systems~\cite{schollwock2011density, vzunkovivc2016dynamical,vzunkovivc2018dynamical,lerose2018chaotic}. Recently, they have been adopted to machine learning problems, in particular classification~\cite{stoudenmire2016supervised,zunkovic2022Deep,stoudenmire2018learning}, generative modelling~\cite{sun2020generative}, image segmentation~\cite{selvan2021patch}, anomaly detection~\cite{wang2020anomaly}, and rule learning~\cite{zunkovic2022Deep, vzunkovivc2022grokking}. Although TNs provide competitive results, they rarely achieve state-of-the-art. A notable example is anomaly detection with tensor networks~\cite{wang2020anomaly}, which is the basis of our approach. Following \cite{wang2020anomaly} we classify the samples to the positive/negative class based on a feature-space distance to the reference positive/negative state. The tightness of the models is ensured by minimizing the Frobenius norms of the reference states. Besides classification, our model enables efficient unbiased sampling of positive and negative samples.

Our model is tested on synthetic point datasets, the MNIST image dataset, and 15 categorical/mixed datasets from the UCI machine learning repository. Presented results are first where a TN model outperforms the state-of-the-art deep neural network (in this case, a generative adversarial network  - GAN) on the image and categorical and mixed datasets. The model is also applicable to missing attributes and in a more general semi-supervised setting with arbitrary number classes. Besides a new TN approach to PUL learning, a new metric is introduced, which applies to model selection and hyperparameter tuning with unlabeled data.

\textbf{Summary of main contributions:}
\begin{itemize}
    \item The anomaly detection model is adapted to the positive unlabeled problem. The modification includes adding a new reference state, modifying the loss terms $\mathcal{L}_{1,2,3}$ to handle labeled and unlabeled data, modifying the Frobenius loss term $\mathcal{L}_4$ to balance the norms of the positive and the negative reference states, and adding a term  $\mathcal{L}_5$ to solve the class collapse problem.
    \item In contrast to \cite{wang2020anomaly}, the introduced TN model enables the generation of new positive and negative samples and has a natural probabilistic interpretation of all loss terms.
    \item A new metric based on the fraction of matching labels between best-performing models is introduced. It is applicable for model selection and hyperparameter tuning with unlabeled data. 
    \item The introduced model significantly improves the positive unlabeled learning state-of-the-art results on the MNIST image and 15 categorical datasets.
\end{itemize}

The paper is organized as follows. Related work and models are reviewed in \sref{sec:related work}. The model is introduced in \sref{sec:model} and the results are presented in \sref{sec:results}. Finally, conclusions and future research directions are discussed in \sref{sec:conclusion}.

\section{Related work}
\label{sec:related work}
Most PUL approaches fall into one of the four categories~\cite{bekker2020learning}: two-step techniques, biased learning, class prior incorporation techniques, and generative adversarial networks. 

The two-step approaches first identify reliable negative samples and then use (semi-)supervised techniques to train a classifier. To perform the first step, we implicitly assume that close samples are labeled similarly. We then identify negative samples as unlabeled samples far from any labeled positive sample. We perform this identification based on a non-traditional classifier or a particular distance metric, e.g., cosine distance and term frequency-inverse document frequency. Different distance measures apply to different data types and are domain-restricted. Most applications have focused on text~\cite{yu2002pebl,xiao2011similarity,zhou2009learning}, and only few on categorical data~\cite{ienco2016positive} or even mixed (categorical and numeric) data~\cite{basile2019ensembles}. As discussed in \sref{sec:model-summary}, our approach has some features of a two-stage approach.

Biased learning techniques treat the unlabeled data as negative and assume that the negative labels have a large amount of noise, i.e., many negative labeled samples are positive~\cite{bekker2020learning}. We learn from such datasets using traditional binary classification methods but with a higher weight on positive samples. Example approaches in this class are weighted SVM-based methods~\cite{elkan2008learning,xiao2011similarity} and probabilistic Latent Semantic Analysis based methods~\cite{zhou2009learning}. Several methods, e.g., \cite{xiao2011similarity}, use a higher weight for positive samples in combination with identifying reliable negative samples (a two-stage technique). The presented approach also uses a higher weight for labeled positive samples incorporating some biased learning techniques (see \sref{sec:model-loss}).
 
The third, class prior techniques, incorporate knowledge about the labeling mechanism. We do this by adjusting a non-traditional classifier by the label frequency (postprocessing)~\cite{elkan2008learning}, weighting the data (by the label frequency), and then training the classifier~\cite{elkan2001foundations}, or changing the learning algorithm to include the label frequency~\cite{gan2017bayesian}. It is possible to incorporate the class prior into the learning objective of the tensor network model. However, this remains an open problem for future research.

Finally, generative adversarial network (GAN) techniques have been adopted for the PUL problem. Most of the GAN approaches are two-stage techniques, where we first train a generator of negative (and sometimes also positive) samples and then use it to train a binary classifier on generated negative and labeled positive data~\cite{hou2017generative,chiaroni2018learning,chiaroni2019generating}. Recently, a single-stage technique~\cite{papivc2023conditional} has been proposed, which simultaneously trains the generator and the classifier network. Despite being a generative model, the presented single-stage tensor network method is much simpler than the GAN approaches, can generate positive and negative samples, and applies to categorical datasets. 

Some positive unlabeled learning methods (e.g., \cite{denis2003text,zhou2012multi}) are based on the idea of co-training~\cite{blum1998combining}. In co-training, we simultaneously train two models on labeled and unlabeled data (typically in a semi-supervised manner). The goal is to obtain two models that predict identical labels. We use a similar idea for model selection and hyperparameter tuning~(see \sref{sec:selection}). The main difference is that we train the models independently and use the label agreement fraction as a proxy metric for accuracy (a co-labeling metric). Incorporating original co-training ideas into the workflow remains for future research.  

We evaluate our approach on the MNIST image and 15 categorical datasets. Accordingly, we use different model sets for comparison/evaluation. We compare the accuracy of the presented model on the MNIST dataset with several GAN approaches:
\begin{itemize}
    \item Generative positive and unlabeled framework (GenPU)~\cite{hou2017generative}: a series of five neural networks: two generators and three discriminators evaluating the positive, negative, and unlabeled distributions.
    \item Positive GAN (PGAN)~\cite{chiaroni2018learning}: Uses the original GAN architecture and assumes that unlabeled data are mostly negatived samples.
    \item Divergent GAN (DGAN)~\cite{chiaroni2019generating}: Standard GAN with the addition of a biased positive-unlabeled risk. DGAN can generate only negative samples.
    \item Conditional generative PU framework (CGenPU)~\cite{papivc2023conditional}: Uses the auxiliary classifier GAN with a new PU loss discriminating positive and negative samples. We use the trained generator network to generate training samples for the final binary classifier (two-stage approach).
    \item Conditional generative PU framework with an auxiliary classifier (CGenPU-AC)~\cite{papivc2023conditional}: The same as CGenPU, but uses an additional classifier to classify the test data (single stage approach).
\end{itemize}

On categorical data, the GAN approaches do not perform well. Therefore, we use the approaches evaluated in \cite{ienco2016positive} and \cite{basile2019ensembles} as baselines:
\begin{itemize}
    \item Positive Naive Bayes (PNB)~\cite{denis2003text,calvo2007learning,bekker2020learning}: calculates the conditional probability for the negative class by using the prior weighted difference between the attribute probability and the conditional attribute probability for the positive class.
    \item Average Positive Naive Bayes (APNB)~\cite{denis2003text,calvo2007learning,bekker2020learning}: Differs from PNB in estimating prior probability for the negative class. PNB uses the unlabeled set directly, while the APNB estimates the uncertainty with a Beta distribution.
    \item Positive Tree Augmented Naive Bayes (PTAN)~\cite{friedman1997bayesian}:  builds on PNB by adding the information about the conditional mutual information between attributes $i$ and $k$ for structure learning.
    \item Average Positive Tree Augmented Naive Bayes (APTAN)~\cite{friedman1997bayesian}:  Differs from PTAN in estimating prior probability for the negative class. PTAN uses the unlabeled set directly, while the APTAN estimates the uncertainty with a Beta distribution.
    \item Laplacian Unit-Hyperplane Classifier (LUHC)~\cite{shao2015laplacian}: biased PUL technique, which determines a decision unit-hyperplane by solving a quadratic programming problem. 
    \item Positive Unlabeled Learning for Categorical datasEt (Pulce) approach~\cite{ienco2016positive}: a two-stage PUL approach that uses a trainable distance measure Distance Learning for Categorical Attributes (DILCA) to find reliable negative samples. In the second stage, we use a k-NN classifier to determine the class.
    \item Generative Positive-Unlabeled (GPU) approach~\cite{basile2019ensembles,bekker2020learning}: Learns a generative model (typically by using probabilistic graphical models (PGMs)) from labeled positive samples. We determine reliable negative samples as the ones with the lowest probability given by the trained generative model. Finally, we train a binary classifier (typically a support vector machine) on labeled positive and reliable negative samples. This approach has also been extended by aggregating many PGMs in an ensemble~\cite{basile2019ensembles}.
\end{itemize}

Our main technical tool is a tensor network called the locally purified state (LPS)~\cite{glasser2019expressive} recently applied to a related anomaly detection problem~\cite{wang2020anomaly}. We use the model in \cite{wang2020anomaly} as a starting point and extend it by considering two tensor networks and adopting the loss for the positive and unlabeled settings. We also introduce an additional loss term necessary to avoid class collapse. Finally, we use the trained models to generate new positive and negative samples, which has not been discussed in~\cite{wang2020anomaly}.

The most common application of tensor networks in machine learning is as a linear model in the exponentially large feature space~\cite{stoudenmire2016supervised,zunkovic2022Deep,stoudenmire2018learning, sun2020generative, selvan2021patch, wang2020anomaly}. Recently a deep tensor network with arbitrary non-linearities has been proposed~\cite{zunkovic2022Deep} and applied study the grokking phenomenon~\cite{power2022grokking, vzunkovivc2022grokking}. Another important application of tensor networks in machine learning is the compression of big deep-network models~\cite{novikov2015tensorizing,panagakis2021tensor}. Finally, tensor networks are used in big data applications~\cite{cichocki2014tensor}, e.g., dimensionality reduction~\cite{cichocki2017tensor}, dynamically weighted directed networks~\cite{luo2021adjusting,chen2022mnl}, and tensor completion~\cite{zheng2021fully}.

\section{Model}
\label{sec:model}
In this section, we will first provide a concise overview of the model, which we shall present in detail in the following two subsections. Then, we will discuss the generation of new positive and negative samples. Finally, we will discuss the model selection and hyperparameter tuning in the positive unlabeled setting.
\subsection{Overview}
\begin{figure}[!htb]
    \centering
    \includegraphics[width=230px]{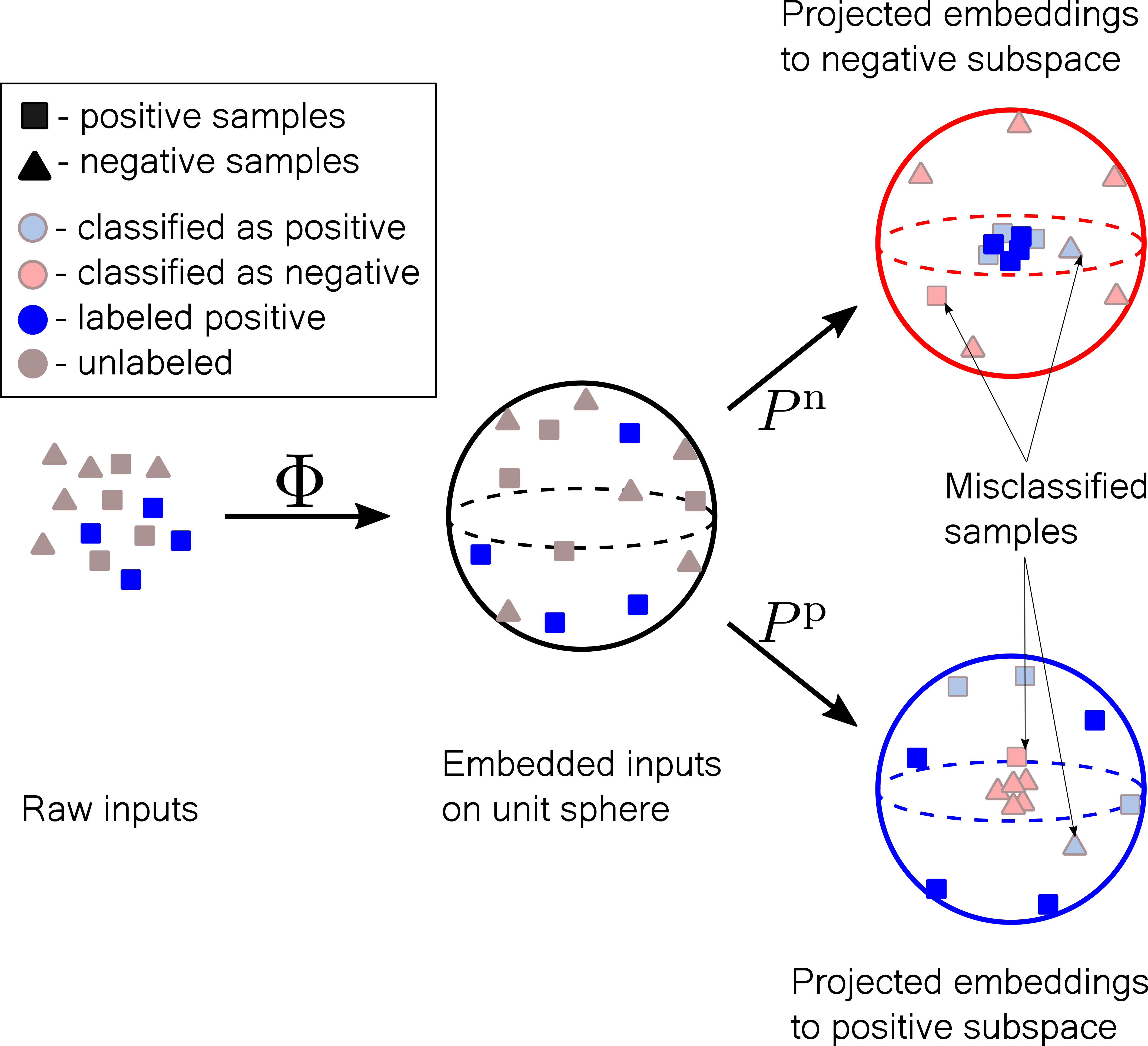}
    \caption{Schematic representation of a trained tensor-network model for positive unlabeled learning. The shape (triangle or square) denotes the actual class, and the color indicates the provided label -- gray (unlabeled) and blue (positive/square).}
    \label{fig:model schematic}
\end{figure}

Our model is a tensor-network kernel method inspired by the tensor-network anomaly detection model \cite{wang2020anomaly}. We show a schematic representation of the model in \fref{fig:model schematic}. We first embed the raw inputs $x$ of size $N$ with an embedding function/layer $\Phi$ onto a unit sphere in an exponentially large vector space $V\subset R^{d^N}$. The dimension $d$ is an embedding parameter. Since the embedded space is high-dimensional, we separate the data by projections $P^{\rm p,n}$ onto positive and negative subspace $W^{\rm p,n}$. Hence, we transform the inputs in two ways as $\hat{y}^{\rm p,n}(x)=P^{\rm p,n}\Phi(x)$. The positive map $\hat{y}^{\rm p}$ ''projects`` positive instances onto a hypersphere of radius $e^{\mu_{\rm 0}}$ centered at the origin. Negative instances have a large overlap with the kernel of the positive map $\hat{y}^{\rm p}$ and are mapped close to the origin (center of the positive hypersphere). Similarly, the negative map $\hat{y}^{\rm n}$ ''projects`` negative instances onto a hypersphere of radius $e^{\mu_{\rm 0}}$ centered at the origin. Positive instances have a large overlap with the kernel of the negative map $P^{\rm n}$ and are mapped close to the origin (center of the negative hypersphere). An instance is recognized as positive if the norm of its positive projection $\hat{y}^{\rm p}$ is larger than the norm of its negative projection $\hat{y}^{\rm p}$, namely $||\hat{y}^{\rm p}(x)||_2>||\hat{y}^{\rm n}(x)||_2$. We denote actual positive samples with $x^{\rm p}$ and actual negative samples with $x^{\rm n}$ and concisely write the action of the model as follows 
\begin{align}
    \label{eq:model summary}
    ||\hat{y}^{\rm p}\left(x^{\rm p}\right)||_2 \approx e^{\mu_0} \gg ||\hat{y}^{\rm p}\left(x^{\rm n}\right)||_2\approx 0, \\ \nonumber
    ||\hat{y}^{\rm n}\left(x^{\rm p}\right)||_2 \approx 0 \ll ||\hat{y}^{\rm n}\left(x^{\rm n}\right)||_2 \approx e^{\mu_0}.
\end{align}
The positive and negative subspaces $W^{\rm p,n}$ are still exponentially large. Yet, their dimension is much smaller than the dimension of the embedding vector space, $\dim(W^{\rm p,n})\ll \dim(V)=d^N$. We map to lower dimensional spaces to ensure that the kernels of $\hat{y}^{\rm p}$ and $\hat{y}^{\rm n}$ are sufficiently large to contain the corresponding negative and positive distributions, respectively.

\subsection{Tensor network architecture} 
\label{sec:TN architecture}
We use tensor networks to make the norm calculation in \eref{eq:model summary} tractable. As shown in \fref{fig:model schematic}, the total model consists of one embedding layer and two projectors $P^{\rm p.n}$, which we represent by LPS tensor networks. We will first discuss the embedding, then the projectors. Finally, we will explain the generation of new positive and negative samples.
\paragraph{Embedding} We will assume that the inputs $x$ are real vectors of size $N$ with elements in the unit interval. If one or more attributes of the input do not have the correct form, we transform them in the preprocessing step. We define the embedding with a local vector transformation $\phi$, which maps each input vector element into a local vector space of dimension $d$. We use the following one-parameter Fourier embedding
\begin{align}
    \label{eq:local embedding}
    \phi(x_i)=(1,\cos(x_i\pi),\ldots,\cos((d-1) x_i\pi)),
\end{align}
where the parameter $d$ determines the dimension of the embedding vector space. The complete embedding transformation $\Phi$ is then given by the tensor product of local embeddings
\begin{align}
    \label{eq:full embedding}
    \Phi(x)=\phi(x_1)\otimes\phi(x_2)\otimes\ldots \phi(x_N).
\end{align}
We will show the tensor-network calculations in a diagrammatic notation \cite{stoudenmire2016supervised,stoudenmire2018learning}. In this notation, we represent the local and the full embedding as
\begin{align}
    \phi(x_i)&=\vcenter{\hbox{\includegraphics[width=20px]{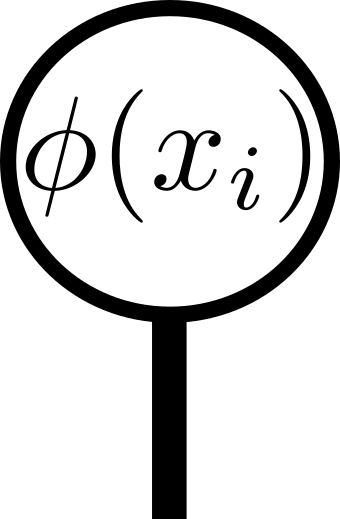}}},\\ \nonumber
    \Phi(x)&=\vcenter{\hbox{\includegraphics[width=130px]{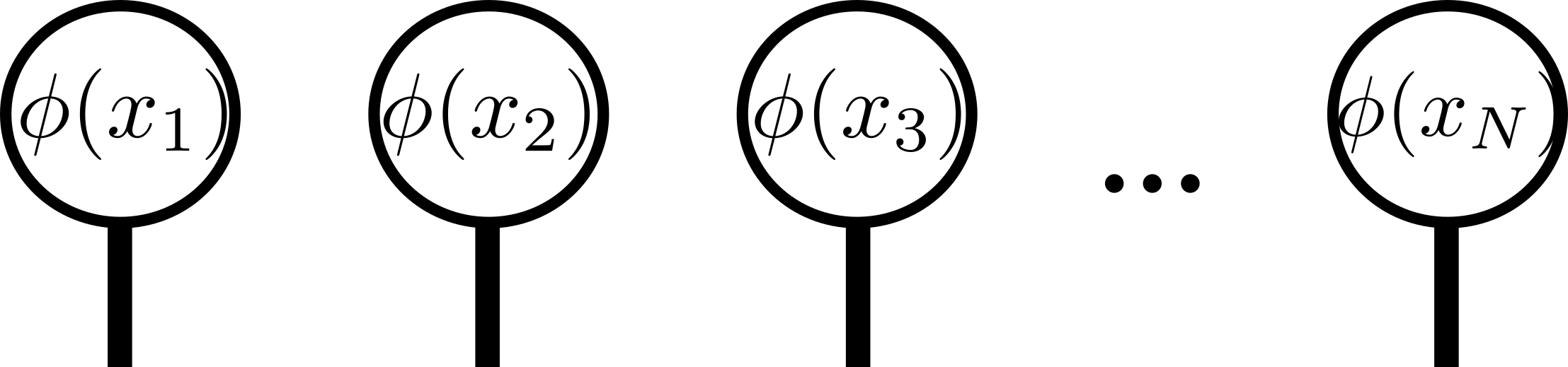}}}~.
\end{align}
Instead of the cosine, we sometimes use the sine basis functions. We may use distinct basis functions for different attributes (elements of the input $x$). An essential property of the local embedding functions is the element-wise orthonormality on the domain, i.e., $\int_0^1\dd u \phi_a(u) \phi_b(u)=\delta_{a,b}$, which will enable efficient sampling~\cite{stoudenmire2016supervised}. In the case of an unbounded domain, we apply a finite set of orthonormal functions (on that domain) without changing the properties of the proposed model. We interpret the embedding functions as basis functions encoding (expanding) the data probability distribution and determine the expansion coefficients by the tensor-network projectors $P^{\rm p,n}$ discussed in the next section.

\paragraph{Tensor-network projector} We use two separate LPS tensor networks representing the positive/negative projector $P^{\rm p,n}$. An LPS is a one-dimensional tensor network consisting of three-dimensional parameter tensors $A^i\in\mathds{R}^{D\times D\times d}$ and four-dimensional parameter tensors $B^i\in\mathds{R}^{D\times D\times d\times d}$. It represents an exponentially large non-square real matrix which we diagrammatically write as
\begin{align}
    \vcenter{\hbox{\includegraphics[width=200px]{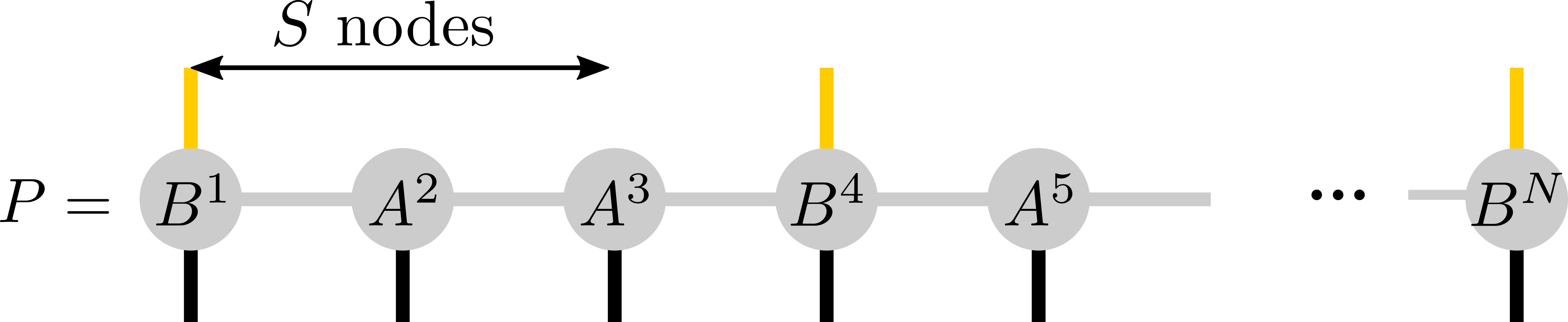}}}~,
\end{align}
where $S$ is a hyper-parameter of the LPS. The bottom legs act on the input space $V$, and the upper legs act on the output space $W$. The number of the top legs is $S$-times smaller than the number of the bottom legs, which guarantees that $\dim(V)$ is $d^S$ times smaller $\dim(W)$. Exponentially smaller output space ensures that the kernel is sufficiently large to contain the data distribution of the opposite class. We calculate the norm of $\hat{y}$ associated with an LPS by contracting the following tensor network
\begin{align}
    \label{eq:y contraction}
    ||\hat{y}(x)||_2^2 = \vcenter{\hbox{\includegraphics[width=180px]{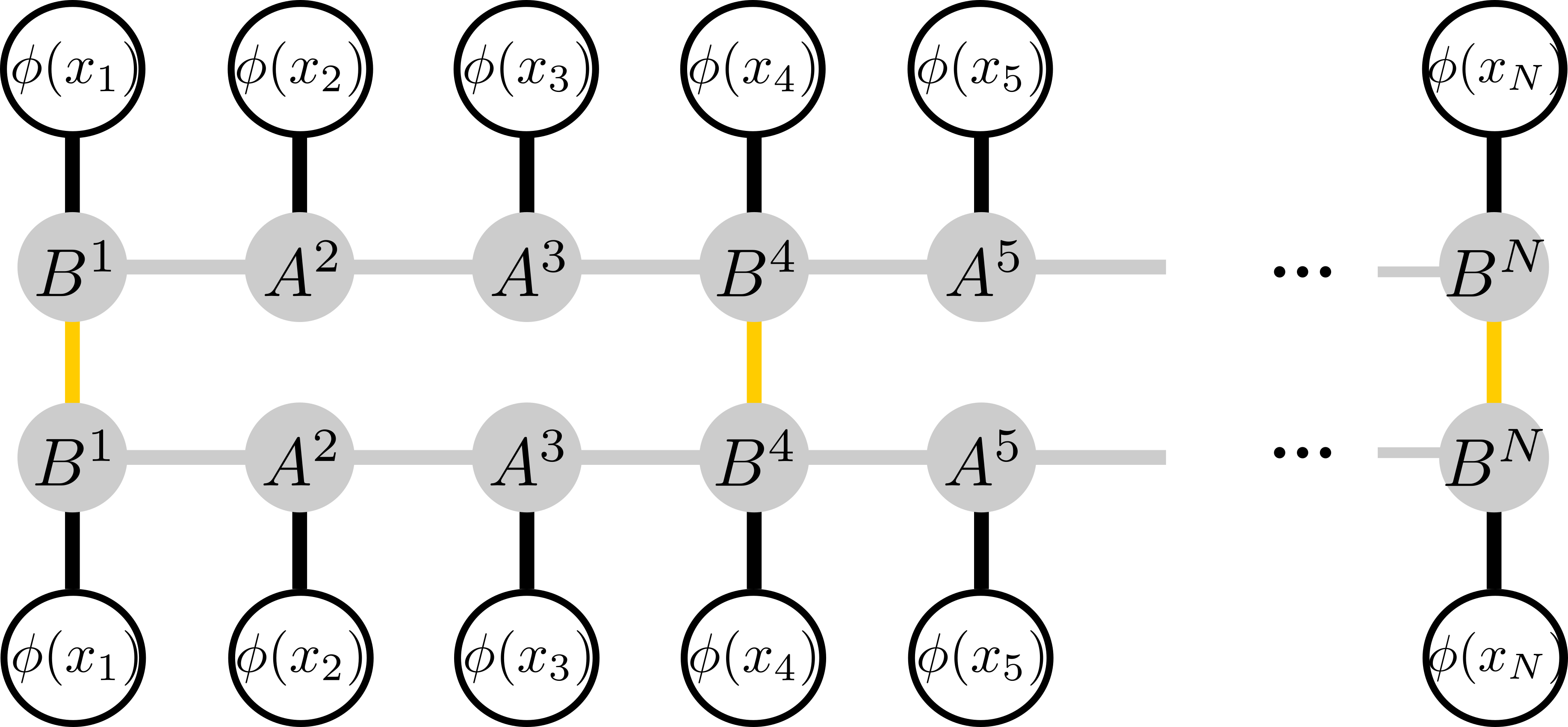}}}.
\end{align}
\paragraph{Contraction complexity} We perform the contraction of \eref{eq:y contraction} efficiently in $O(ND^2(D+d)(\frac{d}{S}+1))$ operations (see e.g. \cite{wang2020anomaly}). Besides the maps $\hat{y}^{\rm p,m}$, we also need the Frobenius norm of the projectors $P^{\rm p,n}$. We calculate the Frobenius norm efficiently in $O(ND^2d(\frac{d}{S}+1))$ operations by contract the following tensor network \cite{wang2020anomaly}
\begin{align}
||P||_F^2=\vcenter{\hbox{\includegraphics[width=170px]{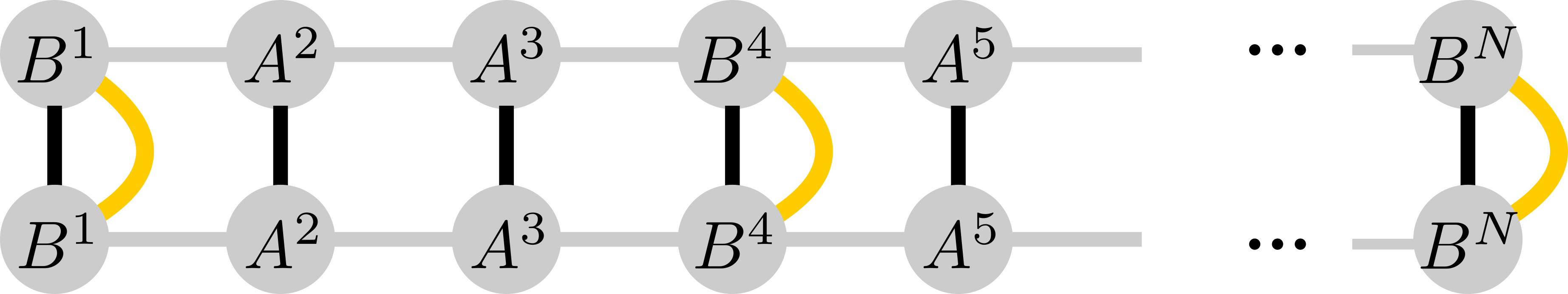}}}~.
\end{align}

\paragraph{Workflow} The entire production workflow of the proposed tensor network model is summarised as follows. 
\begin{enumerate}
    \item {\bf Flatten} If the input is multi-dimensional, we flatten it such that the first dimension is the batch dimension and the second is the feature dimension. We use $N$ to determine the number of features. After flattening the working tensor if of size $n_{\rm batch}\times N$, where $n_{\rm batch}$ determines the batch size.
    \item {\bf Normalization} We normalize the data to the domain of the embedding map. We use an embedding with a unit interval domain for each feature. Hence we normalize all features to the unit interval. Categorical features are first represented as integers and then normalized.
    \item {\bf Embedding} We transform each feature by an embedding function, which may vary for different features. However, we use the same embedding function for all features (see \eref{eq:local embedding}). The embedded tensor has dimension $n_{\rm batch}\times N\times d$, where $d$ is the dimension of the local embedding. 
    \item {\bf Log-norm} Finally, we calculate the norms of positive and negative projections by contracting the embedding tensor with the LPS tensor networks (see \eref{eq:y contraction}). To avoid numerical overflow, we calculate the log-norm of the positive and negative projections. The class is positive if the log norm of the positive projection is larger than the log-norm of the negative projection. 
\end{enumerate}

\subsection{Loss}
\label{sec:model-loss}
We will construct a loss that pushes the positive labeled/classified instances towards the boundary of the positive hypersphere and into the kernel of the negative map $\hat{y}^{\rm n}$ (terms $\mathcal{L}_1$ and $\mathcal{L}_2$) and the opposite for negative classified samples (term $\mathcal{L}_3$). To achieve a tight positive/negative subspace $W^{\rm p,n}$ we will penalize the norm of the matrices $P^{\rm p,n}$ (term $\mathcal{L}_4$). Finally, we will add a term that will prevent the collapse of the classifier to one class (term $\mathcal{L}_5$).

At each stage of training, we separate a data batch into three groups: (i) labeled positive samples $\mathcal{D}^{\rm l}$, (ii) unlabeled samples classified as positive $\mathcal{D}^{\rm p}$, and (iii) unlabeled samples classified as negative $\mathcal{D}^{\rm n}$. For labeled positive samples, i.e. $x\in\mathcal{D}^{\rm l}$, we define the following loss
\begin{align}
    \label{eq:loss1}
    \mathcal{L}^1=\frac{1}{|\mathcal{D}^{\rm l}|}\sum_{x\in\mathcal{D}^{\rm l}}\big(
    &\lambda_1 (\log(||\hat{y}^{\rm p}(x)||_2)-\mu_0)^2 \\ \nonumber
    &+\lambda_2 (\log(||\hat{y}^{\rm n}(x)||_2)-\mu_1)^2
    \big),
\end{align}
where $\lambda_{1,2}$ are loss parameters. We fix $\mu_0=5$ and $\mu_1=-50$. We denote with $|\mathcal{D}|$ the number of elements in the batch/dataset $\mathcal{D}$. The first term of the loss $\mathcal{L}_1$ ensures that the positive projections $\hat{y}^{\rm p}(x)$ have a norm close to $\exp{\mu_0}$. The second term of $\mathcal{L}_1$ pushes the labeled samples towards the kernel of the map $\hat{y}^{n}$. We use the logarithm of the norms to stabilize the training.

The second loss term $\mathcal{L}_2$ concerns the positive classified samples, i.e., $x\in\mathcal{D}^{\rm p}$, which we treat as labeled positive samples. Therefore, we define
\begin{align}
    \label{eq:loss2}
    \mathcal{L}_2=\frac{1}{|\mathcal{D}^{\rm p}|}\sum_{x\in\mathcal{D}^{\rm p}}\big(
    &\lambda_3 (\log(||\hat{y}^{\rm p}(x)||_2)-\mu_0)^2\\ \nonumber
    & +\lambda_4 (\log(||\hat{y}^{\rm n}(x)||_2)-\mu_1)^2
    \big),
\end{align}
with new loss parameters $\lambda_{2,3}$ and the same constants $\mu_{0}=5$ and $\mu_1=-50$.

In the case of negative classified samples, i.e., $x\in\mathcal{D}^{\rm p}$, we reverse the roles of the positive and negative maps $\hat{y}^{\rm p,n}$. The vector $\hat{y}^{n}(x)$ should have a norm close to $\exp(\mu_0)$, while the norm of $\hat{y}^{p}(x)$ should be close to zero. Therefore, we define
\begin{align}
    \label{eq:loss3}
    \mathcal{L}_3=\frac{1}{|\mathcal{D}^{\rm n}|}\sum_{x\in\mathcal{D}^{\rm n}}\big(
    &\lambda_5 (\log(||\hat{y}^{\rm p}(x)||_2)-\mu_0)^2\\ \nonumber 
    &+\lambda_6 (\log(||\hat{y}^{\rm n}(x)||_2)-\mu_1)^2
    \big),
\end{align}
with new loss parameters $\lambda_{5,6}$ and the same constants $\mu_{0,1}$.

To learn a tight fit of the positive and negative distributions we want the Frobenius norms of the projectors $P^{\rm p,n}$ to be as close to one as possible. Moreover, to avoid a collapse to one class, we want to ensure that the Frobenius norms of the projectors are close to each other. We encourage both objectives with the fourth loss term
\begin{align}
    \label{eq:loss4}
    \mathcal{L}_4= \lambda_7\big(&|\log(||P^{\rm p}||_F)|+|\log(||P^{\rm n}||_F)| \\ \nonumber &+ |\log(||P^{\rm p}||_F)-\log(||P^{\rm n}||_F)|\big).
\end{align}

Since the last term does not always prevent a class collapse, we add the fifth term, which ensures the classification of at least some samples in each class. This is done by making the averages of $||\hat{y}^{\rm p,n}||_2$ close to each other, namely
\begin{align}
    \label{eq:loss5}
     \mathcal{L}_5(x)= \lambda_8\Bigg(\frac{1}{|\mathcal{D}^{\rm p}\cup \mathcal{D}^{\rm n}|}\sum_{x\in \mathcal{D}^{\rm p}\cup \mathcal{D}^{\rm n}}&\log(||\hat{y}^{\rm p}(x)||_2) \\ \nonumber
     &- \log(||\hat{y}^{\rm n}(x)||_2)\Bigg)^2.
\end{align}
The parameter $\lambda_8$ is our last hyperparameter.  The entire loss optimized during training is a sum of five different terms 
\begin{align}
    \mathcal{L}=\sum_{i=1}^5\mathcal{L}_i
\end{align}
and has eight hyperparameters. Removing any of the presented loss terms would significantly degrade the performance or lead to a class collapse. 

The positive and negative tensor network projectors determine probability amplitudes (square roots) of the approximate positive and negative distributions (see \sref{sec:sampling}). Therefore, we provide an intuitive probabilistic interpretation of the proposed loss. The first summand in $\mathcal{L}_{1}$ proportional to $\lambda_1$ minimizes the Kullback-Leibler (KL) divergence between the positive-labeled data distribution and the (positive) tensor-network approximation $\hat{y}^{\rm p}$. In contrast, the second summand in $\mathcal{L}_1$ (proportional to $\lambda_2$) maximizes the KL divergence between the positive-labeled data and the (negative) tensor-network approximation $\hat{y}^{\rm n}$. Similar interpretations have the second and third loss term $\mathcal{L}_2$ and $\mathcal{L}_3$: we increase the KL divergence between the opposite-class distributions and decrease the KL divergence between the same-class distributions. The fourth term normalizes the positive/negative tensor-network distributions and captures the remaining parts of the KL divergence in the losses $\mathcal{L}_{1-3}$. The final, fifth loss $\mathcal{L}_5$ ensures that the positive and negative tensor-network distributions have the same distance (KL divergence) to the uniform distribution on the unlabeled data. We can incorporate prior information into our objective by changing the uniform prior in the fifth loss $\mathcal{L}_5$ with another distribution. 

As described in the \sref{sec:results}, we fix the loss hyperparameters once and then keep them constant for all experiments presented in the paper. We find that the best parameter set has a higher weight for labeled samples than the positive/negative classified samples akin to the biased learning techniques discussed in \sref{sec:related work}.

\subsection{Sampling}
\label{sec:sampling}
We will now reformulate the first part of the loss terms $\mathcal{L}_{1,2,3}$ and show that they minimize a distance between the data distribution and positive/negative ``quantum'' probabilities given by $\rho^{\rm p,n}=(P^{\rm p,n})^TP^{\rm p,n}$. We represent a dataset $\mathcal{D}$ as an unnormalized ``quantum'' probability by using the diagonal ensemble
\begin{align}
    \rho(\mathcal{D})=\sum_{x\in\mathcal{D}}\Phi(x)\Phi(x)^T,
\end{align}
where $\Phi(x)$ denotes a column vector embedding and $\Phi(x)^T$ a row vector embedding. We now promote our embedding vector space to a Hilbert space with a Hilbert-Schmidt inner product and interpret the terms in the loss-functions $\mathcal{L}_{1,2,3}$ as the inner product between the actual data distribution and the modeled distributions $\rho^{\rm p,n}$
\begin{align}
    \sum_{x\in\mathcal{D}}\hat{y}^{\rm p,n}(x)=\tr{\left(\rho(\mathcal{D})\rho^{\rm p,n}\right)}
\end{align}
Therefore, the final matrices $P^{\rm p,n}$ define the ``square roots'' of the approximate positive/negative data distributions. 

A distinct feature of many tensor-network models is that they enable efficient unbiased sampling \cite{stoudenmire2016supervised,sun2020generative}, which is also true for probability distributions determined by LPS states. In a tensor network, we typically sample sequentially by constructing a local probability density. Let us assume that we have already sampled all positions until $i+1$. In other words, we know the values $x_1, x_2,\ldots ,x_i$ while the values for positions $i+1, i+2,i+3,\ldots N$ are unknown (to be determined). We sample the attribute at position $i+1$ from the probability density at position $i+1$ given by
\begin{align}
    p(u) = \vcenter{\hbox{\includegraphics[width=190px]{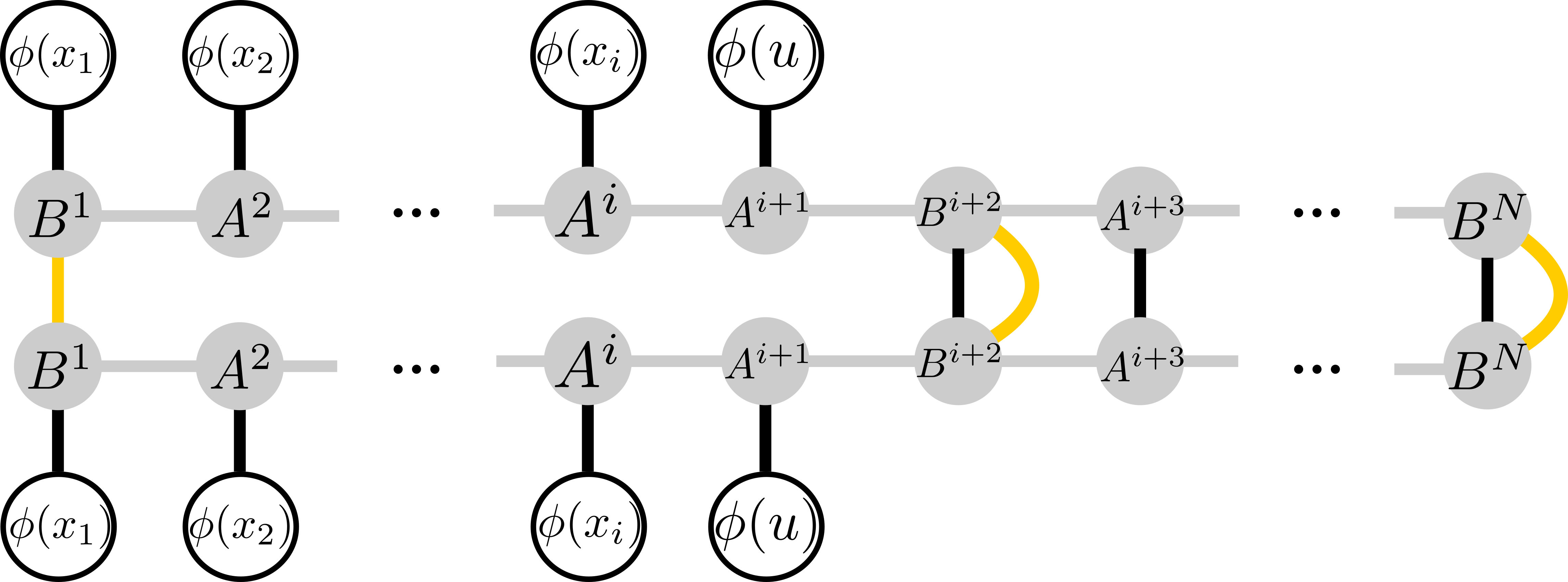}}}~.
    \label{eq:unnormalised p}
\end{align}
Due to the orthonormality of the embedding functions, we replace the marginalization integrals or sums over the variables $x_{i+2},x_{i+3},\ldots x_{N}$ by a simple contraction. The tensor network in \eref{eq:unnormalised p} can be efficiently evaluated and represents an embedding basis function expansion of the unnormalized probability density of the current variable $u$. After we sample the current position, we continue with the next one to the right.

\paragraph{Ensemble sampling} The linearity of the model also enables efficient ensemble sampling. At a given position, we first construct the probability densities corresponding to each LPS in the ensemble and then sample according to the average probability density. Finally, we assign the sampled value to each LPS in the ensemble.

\paragraph{Missing attributes} The LPS model also naturally processes samples with missing attributes. We handle missing attributes by calculating the overlap with the approximate marginal probability distribution, where we marginalize over missing attributes. Specifically, if one or more attributes are missing, we contract the corresponding top and bottom LPS tensors analogously to the normalization \eref{eq:unnormalised p}. We can apply the marginalization over the missing attributes during the training and the prediction phase.

\subsection{Model selection and parameter tuning}
\label{sec:selection}
Before calculating the evaluation metric (accuracy or F1-score), we perform a model selection step. After training several models, we compare the label predictions on the training dataset for each model pair and select the models which agree most. We find that the agreement fraction is a reliable estimate of the final test accuracy of the models (see \sref{sec:results-mnist}). We calculate the evaluation metric only for selected models. Further, we showed that the agreement fraction is also a good metric for hyperparameter tuning (see \sref{sec:results-categorical}) in the case of unlabeled data. Accordingly, we use the same strategy to determine the remaining model hyperparameters: bond dimension $D$, embedding dimension $d$, number of training epochs, and learning rate schedule.

\subsection{Summary} 
\label{sec:model-summary}
We introduce a one-stage tensor-network generative model for positive unlabeled learning. Our model is inspired by the tensor-network anomaly detection model \cite{wang2020anomaly}, but has several key differences. First, we add a second projector that approximates the distribution of negative samples. Second, besides the loss for labeled samples $\mathcal{L}_1$, we add a loss for probably-positive $\mathcal{L}_2$ and probably-negative $\mathcal{L}_3$ samples as well as the ``convergence'' loss $\mathcal{L}^5$, which ensures that the model does not collapse to one class. Third, we present an efficient unbiased scheme for generating new positive and negative samples and show how to process samples with missing attributes. Finally, we introduce a novel model selection and parameter-tuning strategy applicable to unlabeled data.

The proposed setup is similar to a two-stage PUL approach since we use only reliable negative and positive samples for training a classifier. The difference is that we use the same model as a classifier and a generator. Therefore, we omit the second stage, similar to the CGenPU-AC approach. 

\section{Results}
\label{sec:results}
Our approach was tested on three synthetic point datasets, the MNIST image dataset, and 15 categorical/mixed datasets. In all cases, the Adam optimizer was used. The loss parameters were fixed to $\lambda_1=\lambda_2=\lambda_8=4$, $\lambda_4=\lambda_5=2$, and $\lambda_3=\lambda_6=1$. The loss hyperparameters were determined by manual tuning on the point datasets. The exception is the hyperparameter $\lambda_7$, which was determined dynamically after each training epoch. If all labeled samples in the epoch were correctly classified $\lambda_7$ was increased by a factor $k_{\rm inc.}=1.1$ up to the maximum value fixed to 10. In contrast, if the accuracy of the labeled samples in the epoch was smaller than 0.95, $\lambda_7$ was decreased by a factor $k_{\rm dec.}=0.9$ up the minimal value fixed to 0.1. Finally, the factors $k_{\rm inc./dec.}$ were updated at each change from a decrease to an increase of the hyperparameter $\lambda_7$ by setting their new values to $(k_{\rm inc./dec.})^{0.8}$. The hyperparameters of the $\lambda_7$ dynamics were fixed by observing the training on toy datasets. The training dynamics do not significantly depend on the hyperparameter choice. Similarly, the particular choice of the constants $\mu_0=5$, and $\mu_1=-50$ does not significantly change the results on toy datasets as long as the values are aligned with our objective, namely $\mu_0>\mu_1$. 

\subsection{Synthetic datasets}
\label{sec:results-toy}
\begin{figure}[!htb]
    \centering
    \includegraphics[width=0.16\textwidth]{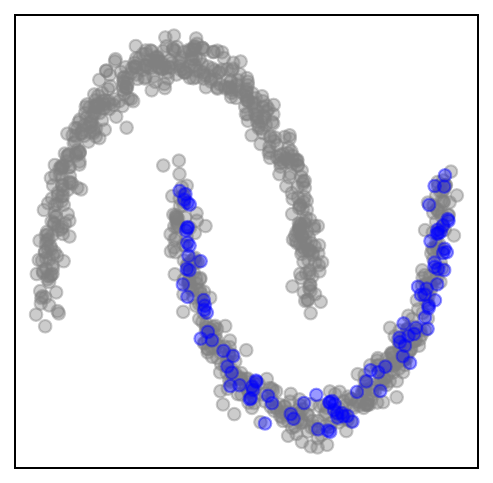}
    \includegraphics[width=0.156\textwidth]{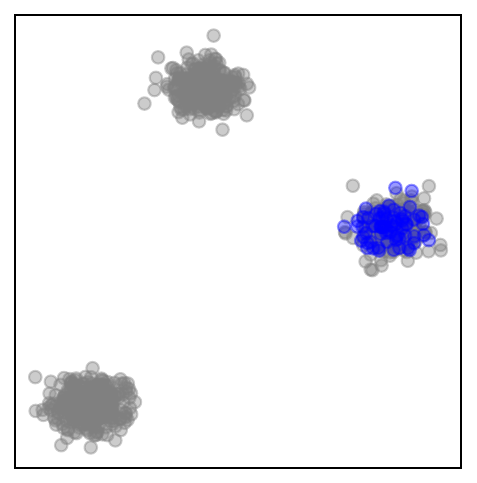}
    \includegraphics[width=0.16\textwidth]{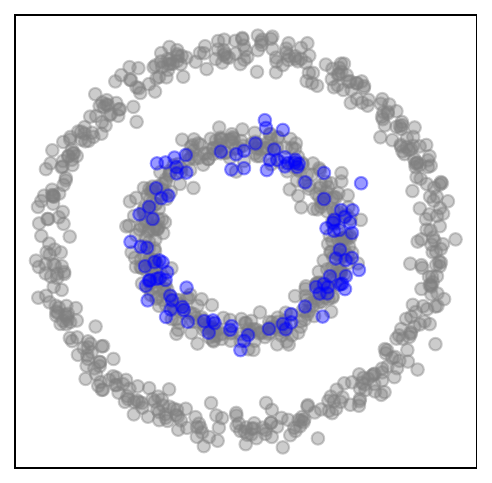}
    
    \includegraphics[width=0.16\textwidth]{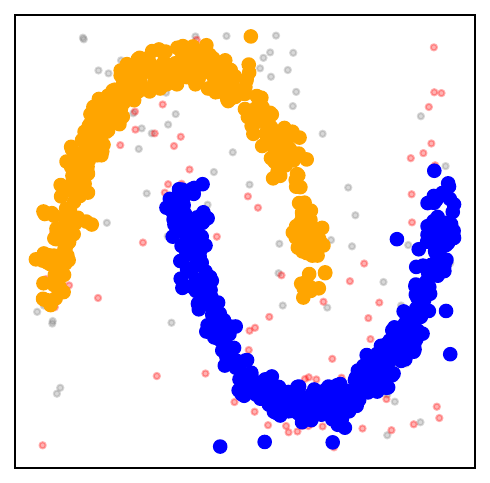}
    \includegraphics[width=0.16\textwidth]{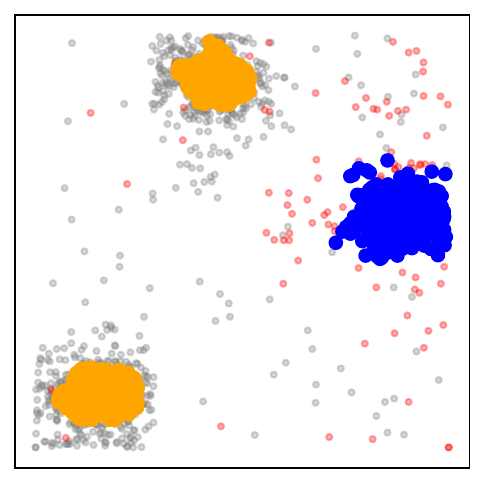}
    \includegraphics[width=0.157\textwidth]{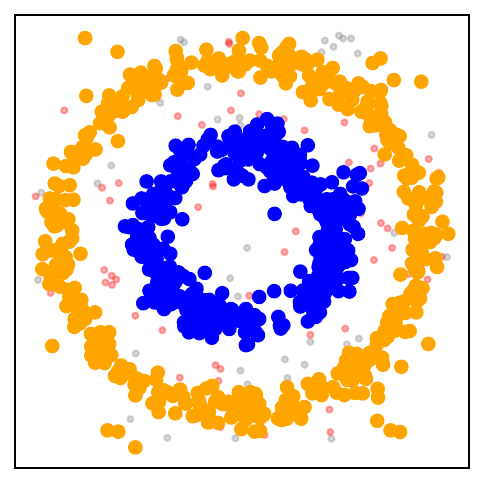}
    \caption{Samples were generated by an ensemble of four models. We show the two moon dataset (left), blobs dataset (middle), and circles dataset (right). The top panels show the training dataset, where gray denotes the unlabeled samples and blue the labeled positive samples. The bottom panels show the generated samples, where orange circles are accepted positive samples, blue circles are accepted-negative samples, and gray and red dots are rejected negative and rejected positive samples.}
    \label{fig:samples}
\end{figure}

The three point datasets, namely the two moon dataset, the circles dataset, and the blobs dataset available through the \textit{scipy} API were used to test the generative properties of our approach. In all cases, 1000 training samples were used, where half of the samples were positive, of which 100 were labeled. To obtain a more expressive model, the input was first repeated nine times, resulting in 18-dimensional feature vectors that were processed as discussed in \sref{sec:model}. Local basis functions were randomly chosen between sine and cosine, and $S=3$, and $D=d=12$ were used. The models were trained with the Adam optimizer and a learning rate of 0.1. After training, sampling from an ensemble was performed by sequential sampling from the average local probability density. Then, samples were accepted only if all models predicted a high probability that the sample is in the correct class. For positive samples, the threshold was set to $\hat{y}^{\rm p}-\hat{y}^{\rm n}>20$. The sampled distribution was further improved by thresholding and removing obvious outliers (see \fref{fig:samples}), although the initial samples were already close to the original distribution. Visually, it is clear that our method better reproduces the original distribution (especially after thresholding) when compared to the state-of-the-art GAN results~\cite{hou2017generative, papivc2023conditional}.

\subsection{MNIST}
\label{sec:results-mnist}
On the image MNIST dataset, the one-vs-one classification for each class pair and the one-vs-all classification for each of the ten classes were performed. The Adam optimizer was used with a learning rate of 0.01 and a batch size of 256. The images were cropped to size $20\times20$ pixels, and random rotation (for the angle $0.05\pi$) and random zoom (with a factor in the range $[0.8,1.2]$) were applied as a data augmentation step. Balanced datasets were used for training and testing on both tasks. Experiments were conducted with $N_{\rm p}=100,10,1$ labeled positive samples. The model parameters were determined by adopting the setting of the previous studies of the TN classification on the MNIST dataset~\cite{wang2020anomaly,zunkovic2022Deep} and fixed to $S=10$, $d=6$, and $D=20$.

\begin{figure}[!htb]
    \centering
    \includegraphics[width=200px]{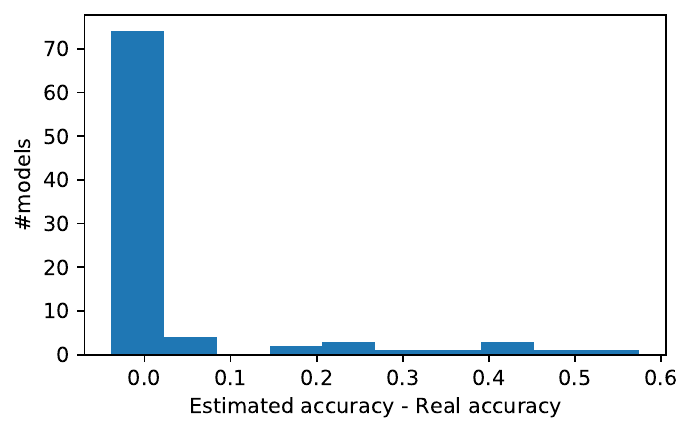}
    \includegraphics[width=200px]{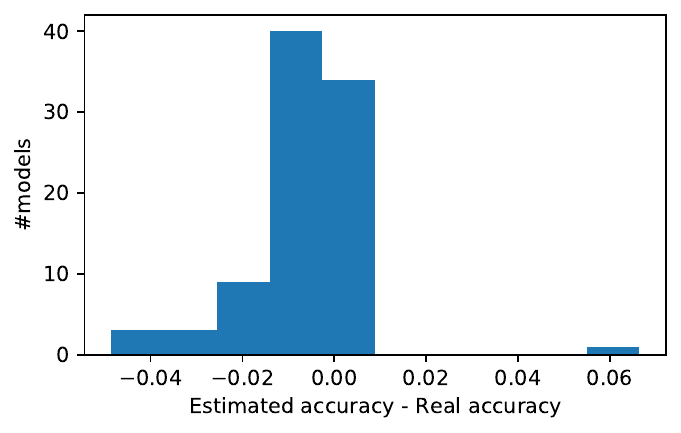}
    \includegraphics[width=200px]{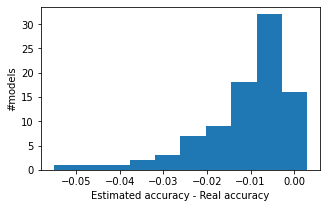}
    \caption{Histogram of differences between the estimated accuracy and the real test accuracy on the one-vs-one MNIST task. From top to bottom we show results for $N_{\rm p}=1,10,100$.}
    \label{fig:corr-acc}
\end{figure}

\paragraph{Batch generation} In a typical PUL scenario, the number of labeled positive samples is much smaller than that of unlabeled samples. Hence, it is likely that many batches will not have any labeled samples during training. This problem is solved by adding a subsample (with repetition) of labeled positive samples. The number of added labeled samples equals the number of samples in the original batch, which contained randomly selected labeled and unlabeled samples.
\paragraph{Model selection and accuracy estimate}
\label{app:corr-acc}
As discussed in \sref{sec:selection}, the model selection has been performed based on the fraction of matching labels. From several trained models, the ones with the largest overlap of predicted labels were evaluated. We refer to the fraction of the matching labels as the estimated accuracy. The effectiveness of this metric on the MNIST dataset is established by comparison with the actual test accuracy. In \fref{fig:corr-acc}, we show the histogram of the difference between the estimated and the test accuracy of the best/selected models on the one-vs-one task. The total number of models equals the total number of class pairs. We perform the comparison for different numbers of labeled samples $N_{\rm p}=1,10,100$. As expected, the estimated accuracy is closer to the test accuracy for more labeled training samples $N_p$. Interestingly, in most cases, the estimated accuracy underestimates the test accuracy. We also observe that for $N_{\rm p}=1,10$ more than 90\% of the differences in \fref{fig:corr-acc} are within one standard deviation of the average test accuracy for the corresponding number of labeled samples $N_{\rm p}$ reported in Table~\ref{tab:one-vs-one mnist}. In the case $N_{\rm p}=100$, approximately 50\% of the differences are within one standard deviation of the average test accuracy reported in Table~\ref{tab:one-vs-one mnist}. However, in this case, the standard deviation is small, namely 0.01, and for all class pairs, the estimated accuracy underestimates the actual test accuracy. 

\paragraph{Comparison with GAN approaches}
As discussed in the introduction, we evaluate the performance of the TN models on the image datasets by comparing the results with the results of the state-of-the-art GAN approaches. We follow \cite{papivc2023conditional} and evaluate the model two tasks. In the one-vs-one task, we select two classes and use one as positive and the other as negative. We repeat this for all possible pairs and then report the average metric. In the one-vs-all task, we select one class as positive and all remaining classes as negative. In both cases, we balance the train and the test datasets to contain the same number of positive and negative samples. Since we balance the datasets, we use the accuracy to evaluate the models. We report the results for the one-vs-one classification task in Table~\ref{tab:one-vs-one mnist} and the results for the one-vs-rest classification task in Table~\ref{tab:one-vs-rest}. Our model performs significantly better than the GAN approaches~\cite{papivc2023conditional}. Notably, in the one-vs-one setting, our approach with only one labeled sample is better than all presented GAN approaches with 100 labeled samples. On the one-vs-rest task, we compare our model only with the CGenPU-AC, PGAN, DGAN, and GenPU approaches~\cite{papivc2023conditional}. Also, in this setting, our approach restricted to only one labeled sample is, on average, comparable to the CGenPU-AC model with 50 labeled samples, which is the state-of-the-art GAN approach. Further, if we increase the number of labeled samples in the one-vs-all setting to 10, we significantly improve the state-of-the-art CGenPU-AC results trained on 50 labeled samples.
\begin{table*}[!htb]
    \centering
    \caption{Average test accuracy on the one-vs-one classification task on the MNIST dataset. A comparison of the GAN approaches and the proposed TN approach. The data for GAN approaches are taken from \cite{papivc2023conditional}.}
    \begin{tabular}{c|cccccc}
        \toprule
         $N_p$ &  PGAN~\cite{chiaroni2018learning} & D-GAN~\cite{chiaroni2019generating}  & GenPU~\cite{hou2017generative}  & CGenPU~\cite{papivc2023conditional}  & CGenPU-AC~\cite{papivc2023conditional} & TN \\
         \midrule
         100 & 0.69 ± 0.23  & 0.77 ± 0.22  & 0.86 ± 0.12 & 0.87 ± 0.19  & 0.89 ± 0.18 & ${\bf 0.99\pm0.01}$\\
         10 &  0.60 ± 0.22 & 0.60 ± 0.22  &   0.59 ± 0.12 & 0.79 ± 0.21   & 0.84 ± 0.18  & ${\bf 0.98\pm0.04}$\\
         1 & 0.84 ± 0.18  & 0.84 ± 0.18  &   0.59 ± 0.14 & 0.71 ± 0.25   & 0.72 ± 0.23  &  ${\bf 0.90\pm 0.20}$\\
         \bottomrule
    \end{tabular}
    \label{tab:one-vs-one mnist}
\end{table*}
\begin{table*}[!htb]
  \caption{Average test accuracy on the one-vs-rest task on the MNIST dataset. The largest accuracy is in bold. The data for GAN approaches are taken from \cite{papivc2023conditional}.}
  \label{tab:one-vs-rest}
  \centering
  \begin{tabular}{c|c|llllllllll|c}
    \toprule
 &$N_{\rm p}$ & 0 & 1 & 2 & 3 & 4 & 5 & 6 & 7 & 8 & 9 & Avg.\\
    \midrule
    \multirow{3}{*}{TN} &100 & {\bf 0.99} & 0.93 & 0.90 & {\bf 0.90} & {\bf 0.93} & {\bf 0.91}  & {\bf 0.98} & 0.90 & {\bf 0.91} & {\bf 0.88} & {\bf 0.92}\\
  &10 & 0.98 & 0.95  & {\bf 0.92} & {\bf 0.90} & 0.91 & 0.83 & {\bf 0.98} & 0.92  & 0.80  & 0.87 & 0.91\\
  &1 & 0.98 & 0.96 & 0.89  & 0.83 &  0.91  & 0.65 & {\bf 0.98} & 0.85 & 0.57 & 0.85 & 0.85 \\
     \midrule
     CGenPU-AC~\cite{papivc2023conditional} &50 & 0.89 & {\bf 0.97} & 0.78 & 0.78 & {\bf 0.93} & 0.80 & 0.87 & {\bf 0.93} & 0.80 & 0.83 & 0.86 \\
     PGAN~\cite{chiaroni2018learning} & 50& 0.72 & 0.63  & 0.65  & 0.64 & 0.66  & 0.62  & 0.66 & 0.63 & 0.56 & 0.64 & 0.64 \\
     D-GAN~\cite{chiaroni2019generating}  & 50 & 0.73 & 0.66 & 0.62 & 0.69 & 0.65 & 0.57 & 0.70 & 0.63 & 0.57 & 0.67 & 0.65 \\
     GenPU~\cite{hou2017generative} & 50 & 0.86 & 0.56 & 0.78 & 0.67 & 0.75 & 0.74 & 0.80 & 0.78 & 0.68 & 0.71 & 0.73 \\ 
    \bottomrule
  \end{tabular}
\end{table*}
\paragraph{Statistical tests}
Since the baseline data for the one-vs-one task is unavailable, we perform the statistical analysis only for the one-vs-rest MNIST task. However, considering the average test accuracy, our improvement over the CGenPU-AC model~\cite{papivc2023conditional} is larger than the improvement of the CGenPu-AC over the GenPU model. Further, our model reduces the error from 1.6 ($N_{\rm p}=1$) to 11 ($N_{\rm p}=100$) times compared with the previous state-of-the-art. Finally, the average accuracy difference between the CGenPU-AC model and our model is larger than four standard deviations, except in the extreme case of $N_{\rm p}=1$ (see Table~\ref{tab:one-vs-one mnist}).

On the one-vs-rest task, we assess the statistical significance of the results on MNIST data by using the Friedman statistics and the Nemenyi test \cite{demvsar2006statistical}. 

In the Friedman test, the null hypothesis is that all the methods perform similarly, i.e., the $\chi^2_{\rm F}$ value is similar to the critical value for the chi-square distribution with $k-1$ degrees of freedom. For the data reported in Table~\ref{tab:one-vs-rest} the Friedman coefficient is $\chi_{\rm F}^2=33.7$ with $\mathrm{p-value}=9\cdot 10^{-7}$. Therefore, we reject the null hypothesis of the Friedman test. 

\begin{figure}[!htb]
    \centering
    \includegraphics[width=\columnwidth]{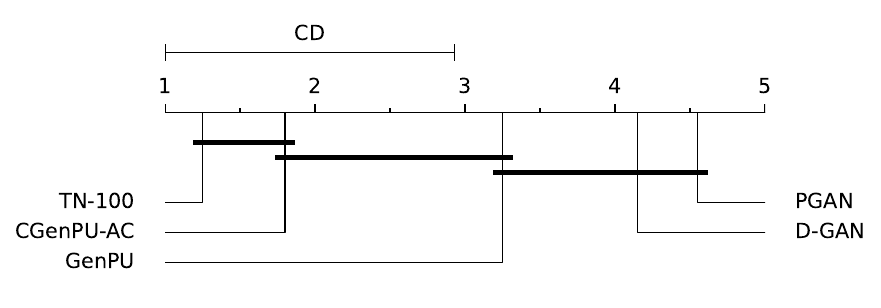}
    \caption{Critical difference graph of the Nemenyi test for the MNIST dataset computed using average rankings of methods and 0.05 significance level.}
    \label{fig:mnist nemenyi}
\end{figure}

In the posthoc analysis, we use the Nemenyi test, which compares the classifier average-rank differences with the critical distance 
\begin{align}
    CD = q_\alpha\sqrt{\frac{k(k+1)}{6N}},
\end{align}
where $k=5$ is the number of models and $N=10$ is the number of tasks (in our case the one-vs-all task), $\alpha$ is the significance level and $q_\alpha$ is the critical value for the two tailed Nemenyi test~\cite{demvsar2006statistical}. For $q_{\alpha=0.05}=2.727774$ the critical distance is $CD=1.93$. Our method is, therefore, statistically significantly better than all but the CGenPU-AC approach (see \fref{fig:mnist nemenyi} for details). The results differ from \cite{papivc2023conditional} since we used significantly fewer experimental runs.

\subsection{Categorical and mixed dataset}
\label{sec:results-categorical}
In this section, we evaluate our approach to PUL on categorical and mixed datasets. We follow the setup of \cite{ienco2016positive} and consider 15 datasets\footnote{All considered datasets are publicly available on the UCI machine learning repository \href{https://archive.ics.uci.edu/ml/}{https://archive.ics.uci.edu/ml/}~.}. For each dataset, we consider three different PUL tasks related to different fractions of labeled positive samples, namely $30\%$, $40\%$, and $50\%$ of all positive samples. Negative samples and the remaining positive samples comprise the unlabeled samples. We thus evaluate our model on 45 different PUL problems. As in \cite{ienco2016positive}, we assume the class with the most instances as positive and the class with the second most instances as negative. In the multi-class setting, we discard the remaining classes. 

In contrast to the MNIST experiments, the train and test datasets are not balanced. The preferred metric in the case of unlabeled and unbalanced data is the F1-score~\cite{ienco2016positive}, which we use to evaluate different approaches to PUL on categorical datasets.

\paragraph{Training details} The preprocessing of the mixed/categorical datasets includes several steps. First, the categorical attributes were converted to numbers. Then, all attributes were normalized to the unit interval. Finally, the embedding functions discussed in \sref{sec:TN architecture} were applied. The use of discrete basis/embedding functions for categorical attributes was avoided for simplicity. Since considered datasets are small, a single batch training without the repetition of labeled samples was performed. The loss hyperparameters were the same as on the synthetic and MNIST datasets.

\paragraph{Hyperparameter tuning} 
For each dataset, the embedding dimension $d$, the bond dimension $D$, the number of train epochs, and the learning rate schedule were determined by hyperparameter tuning as described in \sref{sec:selection}. In all cases, the hyperparameter tuning has been performed by random search on the following intervals: $d\in\{4,12,20\}$, $D\in\{2,6,12,20\}$, $\rm{epochs}\in\{10,20,50,150,210,400\}$. The patience parameter was tied to the number of epochs (see Table~\ref{tab:dataset-description}). Tuned hyperparameters were the same for all three considered fractions of positive samples. The skip size $S$, and the number of repetitions of the input attributes were determined based on the number of all attributes in the dataset. The main guiding principle for their choice was that the dimension of the kernel and the rank of the projectors should be sufficiently large, as discussed in \sref{sec:TN architecture}. The Table~\ref{tab:dataset-description} shows the dataset characteristics and the chosen hyperparameters. The constant attributes have been removed after the preprocessing steps. Therefore, the number of attributes reported in the table differs from the number of all attributes reported in \cite{ienco2016positive}.
\begin{table*}[!htb]
    \centering
    \caption{Dataset characteristics and hyperparameters. With '/' we denote that the patience parameter is not used, namely the learning rate is constant during training.}
    \label{tab:dataset-description}
    \begin{tabular}{c|c|c|c|c|c|c|c|c|c|c}
         \toprule
         Dataset & Pos. samp. & Neg. samp. & Attr.& Rep. & S & d & D & lr & Epochs & Patience \\
         \midrule
           audiology & 57 & 48 & 40 & 1 & 10 & 12 & 12 & 0.01 & 210 & 70   \\
           breast-cancer & 196 & 81 & 9 & 2 &6 &12  &20 &0.1 &10 &/   \\
           chess & 1669 & 1527 & 36 & 2 & 4 & 20&6 & 0.01&400 & 100   \\
           credit-a& 357 & 296 & 15 & 1 & 5 & 12&12 &0.01 & 50 & /   \\
           dermatology&111 & 71 &34 & 1 & 11 & 4& 2& 0.1& 20& /   \\
           heart-c&160 & 160 & 13 & 2& 4 &12 &12 &0.01 & 50 & /   \\
           hepatitis& 123 & 32 & 19 & 1& 4 &12 &12 &0.01 &210 &70   \\
           iris& 50 & 50 & 4 & 2 & 4 & 4 & 2 & 0.1 & 20 & /  \\
           lymph& 81 & 61& 18 &3 & 4 &12 &6 &0.1 &150 & 50   \\
           mushroom& 3488&2156 &22 &2 &7 &12 &6 &0.01 &150 &50   \\
           nursery&4319 & 4266& 8 &2 & 4& 20& 6& 0.1&400 &100   \\
           pima& 500&268 &8 & 2&4 &12 &12 &0.01 &210 &70   \\
           soybean& 92& 91&21 &1 &7 &20 &6 &0.01 &400 &100   \\
           spambase&2788 &1813 & 57& 2 & 10 &12 &12 &0.01 &210 & 70   \\
           vote&124 &108 &16 &2 &4 & 20& 6&0.1 &400 &100   \\
         \bottomrule
    \end{tabular}
\end{table*}

\paragraph{Estimated accuracy and model selection}
To validate the model agreement fraction (accuracy estimate) as a reliable accuracy estimate and metric for model selection and hyperparameter tuning, we compare the estimated accuracy with the F1 score on the test dataset of the 10-fold cross-validation. The accuracy estimate has been calculated as follows. First, a set of 50 hyperparameter tuples was created. Then, ten models were trained for each tuple and each fold in the cross-validation. Out of the ten trained models, only the best model based on the agreement fraction was chosen for the evaluation of the hyperparameter performance. Therefore, for each of the 50 parameter tuples, 100 models were trained in the hyperparameter tuning step, from which only ten were chosen for evaluation, one for each fold in the cross-validation. The test F1 score has been calculated only for experimental validation of the estimated accuracy metric and is not applicable for hyperparameter tuning due to missing labels.

\begin{figure*}[!htb]
    \centering
    \includegraphics[width=0.22\textwidth]{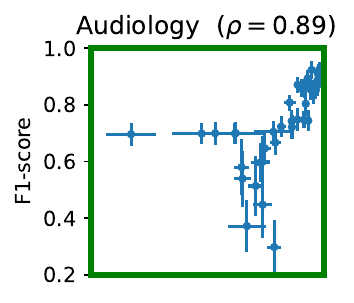}
    \includegraphics[width=0.19\textwidth]{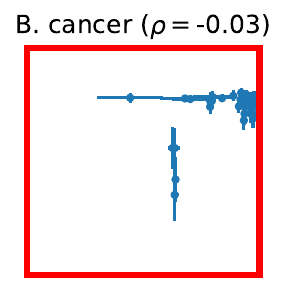}
    \includegraphics[width=0.174\textwidth]{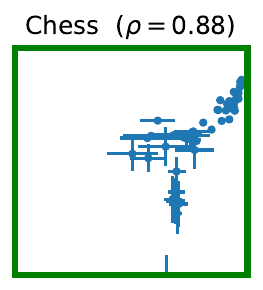}
    \includegraphics[width=0.176\textwidth]{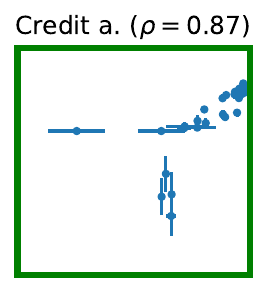}
    \includegraphics[width=0.174\textwidth]{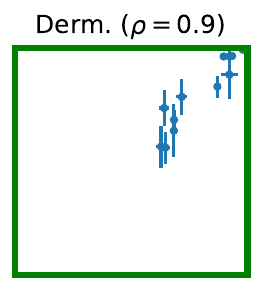}
    
    \includegraphics[width=0.22\textwidth]{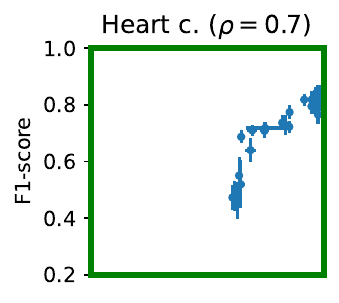}
    \includegraphics[width=0.18\textwidth]{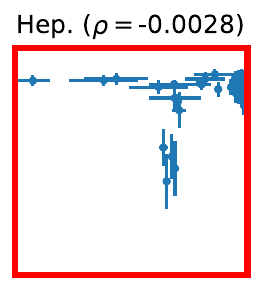}
    \includegraphics[width=0.18\textwidth]{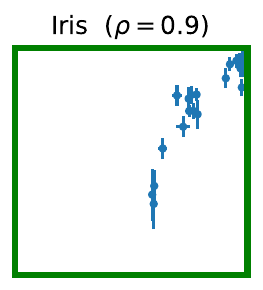}
    \includegraphics[width=0.18\textwidth]{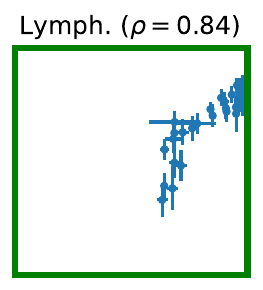}
    \includegraphics[width=0.18\textwidth]{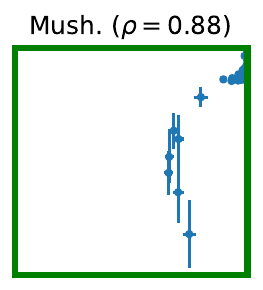}
    
    \includegraphics[width=0.22\textwidth]{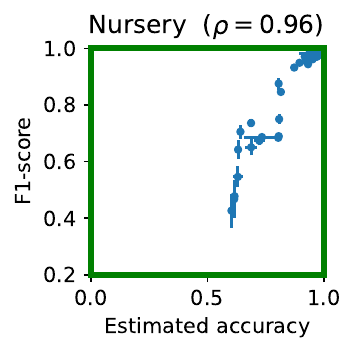}    \includegraphics[width=0.18\textwidth]{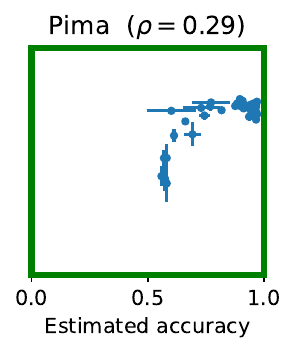}
    \includegraphics[width=0.18\textwidth]{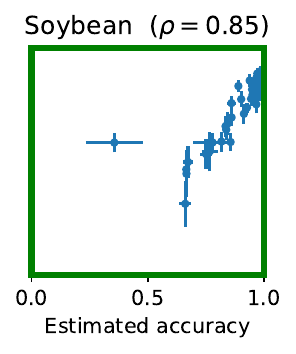}    \includegraphics[width=0.18\textwidth]{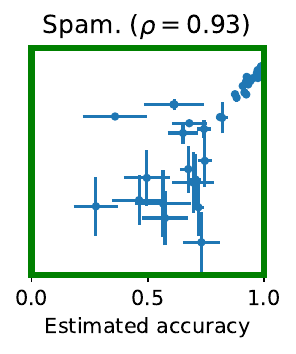}
    \includegraphics[width=0.18\textwidth]{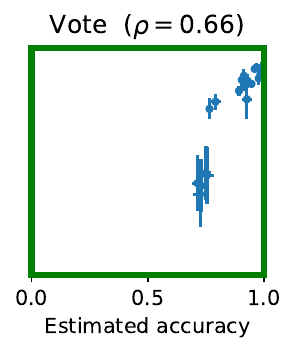}
    \caption{Relation between the mean estimated accuracy and the mean test F1-score. The horizontal and vertical bars represent the standard error of the 10-fold cross-validation. In the title, we report the name of the dataset and the Spearman coefficient $\rho$ for the dataset. The red box indicates if the $p-$value of the Spearman coefficient is over 0.05.}
    \label{fig:corr-acc-categorical}
\end{figure*}

In \fref{fig:corr-acc-categorical}, we show the relation between the estimated accuracy (highest agreement fraction between trained models) and the test F1 score. We observe a monotonic relationship on almost all considered datasets. The only two datasets where monotonic relation is not visually clear are the Cancer and Hepatitis datasets. Even in those datasets, the hyperparameter settings with high estimated accuracy have a high F1 score. To quantify the relation between the estimated accuracy and the test F1 score, we calculate the Spearman coefficient $\rho$ and the corresponding $p$-value. 
From 15 datasets, 13 have significant correlations, i.e., $p-\mbox{value} > 0.05$. By averaging only the significant values, we obtain the Spearman coefficient $\rho=0.81 \pm 0.17$. This experiment demonstrates the utility and reliability of the estimated accuracy for hyperparameter tuning and model selection in case of missing labels. 

\paragraph{Comparison with other methods}
As discussed in the introduction, we compare the results of our TN model with seven different approaches to positive unlabeled learning on categorical datasets. In Table~\ref{tab:categorical f1}, we show the F1-score from the 10-fold cross-validation for each of the considered PUL tasks and models. The results for models PNB, PTAN, APNB, Pulce, and LUHC are taken from \cite{ienco2016positive}\footnote{We do not report the standard deviation for these models since they were not included in the original publication \cite{ienco2016positive}}. The results for the GPU model were reproduced using the code accompanying the paper~\cite{basile2019ensembles}. The GPU model has been reevaluated since it represented the best model for the PUL task on categorical datasets.

In Table~\ref{tab:categorical f1}, we show the F1 score on the test dataset for all dataset-model pairs. We observe that our TN model obtains the best results in 32 out of 45 PUL tasks, which is considerably more than the next-best GPU model with the eight best results. Our model performs worst on hepatitis and Pima datasets, however, on those datasets, it still performs better than the overall second-best GPU model. On the remaining datasets where our model is not the best-performing model, the F1 scores are within one standard deviation of the best-performing model. On datasets chess, heart-c, nursery, spambase, and vote the F1 scores obtained by our model are by more than one standard deviation higher than the next-best model, which is not always the same. On the remaining datasets, our model has the highest F1 score, and the second-best F1 score is within one standard deviation. Finally,
our model also significantly improves the average F1-score from 0.81 to 0.90.
\begin{table*}[!htb]
    \centering
    \caption{Mean F1-score of the 10-fold cross-validation for different PUL classification methods on categorical (and mixed) data. Scores in white columns are reported in \cite{ienco2016positive}, which does not include the standard deviation, and are shown for convenience. The values for the GPU method are reproduced by using the implementation accompanying the paper~\cite{basile2019ensembles}. The last TN column refers to the presented tensor network approach.}
    \label{tab:categorical f1}
\small{
    \begin{tabular}{l c c c c c c c a a}
        \toprule
        {\bf Dataset} & {\bf \%} & {\bf PNB}  & {\bf PTAN} & {\bf APNB}  & {\bf APTAN} &  {\bf Pulce}\footnote{We report the best result of the method \cite{ienco2016positive}}& {\bf LUHC} & {\bf GPU} & {\bf TN} \\
        \midrule
        audiology& 30 & 0.68 & 0.66 & 0.70 & 0.66 & 0.74 & 0.14 & ${0.84\pm 0.04}$ & ${\bf 0.90\pm0.12}$  \\
        audiology& 40 & 0.75 & 0.71 & 0.74 & 0.66 & 0.82 & 0.50 & ${0.88\pm 0.04}$ & ${\bf 0.91\pm0.10}$  \\
        audiology& 50 & 0.80 & 0.78 & 0.80 & 0.71 & 0.89 & 0.70 & ${\bf 0.98\pm 0.04}$ & ${0.94\pm0.10}$  \\[0.1cm]
        
        breast-cancer& 30 & 0.40  & 0.43 & 0.39 & 0.43 & 0.46 & {\bf 0.83} & ${0.48\pm 0.04 }$& ${0.81\pm0.06}$\\
        breast-cancer& 40 & 0.42  & 0.43 & 0.40 & 0.45 & 0.47 & {\bf 0.83} & ${0.48\pm 0.04 }$& ${\bf 0.83\pm0.07}$\\
        breast-cancer& 50 & 0.42  & 0.44 & 0.41 & 0.44 & 0.47 & {\bf 0.83} & ${0.52\pm 0.04 }$& ${\bf 0.83\pm0.06}$\\[0.1cm]
        
        chess& 30 & 0.58 & 0.59 & 0.64 & 0.64 & 0.70 & 0.69 & ${ 0.69 \pm 0.04}$& ${\bf 0.87\pm 0.01}$ \\
        chess& 40 & 0.58 & 0.60 & 0.64 & 0.64 & 0.70 & 0.69 & ${ 0.69 \pm 0.04}$& ${\bf 0.89\pm 0.01}$ \\
        chess& 50 & 0.58 & 0.60 & 0.64 & 0.64 & 0.66 & 0.69 & ${ 0.77 \pm 0.04}$& ${\bf 0.89\pm 0.02}$ \\[0.1cm]
        
        credit-a& 30 & 0.73 & 0.72 & 0.73 & 0.72 & {0.83} & 0.62 & ${0.76\pm 0.10}$ & ${\bf 0.87\pm0.03}$ \\
        credit-a& 40 & 0.73 & 0.72 & 0.72 & 0.72 & {0.85} & 0.62 & ${0.78\pm 0.08}$ & ${\bf 0.87\pm0.04}$ \\
        credit-a& 50 & 0.73 & 0.72 & 0.72 & 0.72 & {\bf 0.87} & 0.62 & ${0.79\pm 0.06}$ & ${\bf 0.87\pm0.03}$ \\[0.1cm]
        
        dermatology& 30 & 0.57 & 0.57 & 0.57 & 0.56 & 0.99 & 0.75 & $ {\bf 1.00\pm 0.00 }$& ${\bf 1.00\pm 0.00}$ \\
        dermatology& 40 & 0.57 & 0.57 & 0.58 & 0.57 & 0.99 & 0.75 & $ {\bf 1.00\pm 0.00 }$& ${\bf 1.00\pm 0.00}$ \\
        dermatology& 50 & 0.59 & 0.57 & 0.60 & 0.58 & 0.99 & 0.75 & $ {0.99\pm 0.04 }$& ${\bf 1.00\pm 0.00}$ \\[0.1cm]
        
        heart-c& 30 & 0.73 & 0.63 & 0.70 & 0.64 & 0.72 & 0.71 & $ {0.77\pm 0.09 }$& ${\bf 0.82 \pm 0.05}$ \\
        heart-c& 40 & 0.77 & 0.63 & 0.78 & 0.70 & 0.76 & 0.71 & $ {0.73\pm 0.10 }$& ${\bf 0.83 \pm 0.07}$ \\
        heart-c& 50 & 0.77 & 0.63 & 0.77 & 0.68 & 0.78 & 0.71 & $ {0.77\pm 0.07 }$& ${\bf 0.83 \pm 0.05}$ \\[0.1cm]
        
        hepatitis& 30 & ${\bf 0.87}$ & 0.85 & ${\bf 0.87}$ & 0.86 & ${\bf 0.87}$ & 0.04 &$ {0.82 \pm 0.04 }$& ${ 0.77 \pm0.04}$ \\
        hepatitis& 40 & ${\bf 0.88}$ & 0.85 & ${\bf 0.88}$ & 0.85 & ${\bf 0.88}$ & 0.04 &$ {0.77 \pm 0.04 }$& ${ 0.80 \pm0.04}$ \\
        hepatitis& 50 & ${\bf 0.88}$ & 0.86 & ${\bf 0.88}$ & 0.85 & ${\bf 0.88}$ & 0.04 &$ {0.76 \pm 0.04 }$& ${ 0.81 \pm0.04}$ \\[0.1cm]
        
        iris& 30 & 0.66 & 0.65 & 0.64 & 0.64 & 0.83 & 0.67 & $ {0.98\pm 0.02 }$& ${\bf 1.00\pm 0.00}$ \\
        iris& 40 & 0.68 & 0.67 & 0.66 & 0.65 & 0.92 & 0.67 & $ {\bf 1.00\pm 0.00 }$& ${\bf 1.00\pm 0.00}$ \\
        iris& 50 & 0.70 & 0.68 & 0.68 & 0.68 & 0.97 & 0.67 & $ {\bf 1.00\pm 0.00 }$& ${\bf 1.00\pm 0.00}$ \\[0.1cm]
        
        lymph& 30 & 0.84 & 0.79 & ${\bf 0.85}$ & 0.84 & 0.81 & 0.0 & $ {0.83 \pm 0.04 }$& ${ 0.80\pm 0.07}$ \\
        lymph& 40 & {\bf 0.84} & 0.79 & ${0.83}$ & 0.81 & 0.74 & 0.60 & $ {0.83 \pm 0.04 }$& ${\bf 0.84\pm 0.07}$ \\
        lymph& 50 & 0.86 & 0.81 & ${\bf 0.87}$ & 0.82 & 0.85 & 0.60 & $ {0.82 \pm 0.04 }$& ${ 0.86\pm 0.07}$ \\[0.1cm]
        
        mushroom& 30 & 0.72 & 0.68 & 0.67 & 0.67 & 0.74 & 0.68 & $ {\bf 0.94 \pm 0.04 }$& ${0.93\pm 0.01}$ \\
        mushroom& 40 & 0.75 & 0.73 & 0.67 & 0.68 & 0.76 & 0.68 & $ {\bf 0.96 \pm 0.04 }$& ${0.94\pm 0.02}$ \\
        mushroom& 50 & 0.75 & 0.74 & 0.68 & 0.69 & 0.82 & 0.68 & $ {\bf 0.98 \pm 0.01 }$& ${0.94\pm 0.01}$ \\[0.1cm]
        
        nursery& 30 & 0.65 & 0.56 & 0.65 & 0.50 & 0.74 & 0.67 & $ {0.81 \pm 0.22 }$& ${\bf 0.99\pm 0.01}$ \\
        nursery& 40 & 0.69 & 0.61 & 0.69 & 0.56 & 0.77 & 0.67 & $ {0.86 \pm 0.15 }$& ${\bf 1.00\pm 0.00}$ \\
        nursery& 50 & 0.69 & 0.74 & 0.70 & 0.44 & 0.81 & 0.67 & $ {0.93 \pm 0.19 }$& ${\bf 1.00\pm 0.00}$ \\[0.1cm]
        
        pima& 30 & 0.49 & 0.50 & 0.50 & 0.50 & 0.52 & {\bf 0.79} & $ {0.59\pm 0.04 }$& ${0.74\pm0.04}$  \\
        pima& 40 & 0.49 & 0.50 & 0.50 & 0.51 & 0.55 & {\bf 0.79} & $ {0.56\pm 0.04 }$& ${0.76\pm0.03}$  \\
        pima& 50 & 0.49 & 0.50 & 0.51 & 0.52 & 0.53 & {\bf 0.79} & $ {0.61\pm 0.04 }$& ${0.77\pm0.03}$  \\[0.1cm]
        
        soybean& 30 & $0.81$ & 0.80 & $0.86$ & 0.81 & 0.76 & 0.67 & $ {0.89 \pm 0.04 }$& ${\bf 0.92\pm0.07}$\\
        soybean& 40 & $0.86$ & 0.84 & $0.86$ & 0.83 & 0.79 & 0.67 & $ {0.88 \pm 0.04 }$& ${\bf 0.92\pm0.08}$\\
         soybean& 50 & $0.92$ & 0.88 & $0.92$ & 0.86 & 0.85 & 0.67 & $ {0.89 \pm 0.04 }$& ${\bf 0.95\pm0.06}$\\[0.1cm]
        
        spambase& 30 & 0.57 & 0.57 & 0.57 & 0.57 & 0.64 & 0.75 & $ {0.77 \pm 0.04 }$& ${\bf 0.93\pm0.01}$\\
        spambase& 40 & 0.57 & 0.57 & 0.57 & 0.57 & 0.66 & 0.75 & $ {0.80 \pm 0.04 }$& ${\bf 0.93\pm0.01}$\\
        spambase& 50 & 0.57 & 0.57 & 0.57 & 0.57 & 0.66 & 0.75 & $ {0.83 \pm 0.04 }$& ${\bf 0.93\pm0.01}$\\[0.1cm]
        
        vote& 30 & 0.62 & 0.56 & 0.62 & 0.55 & 0.68 & 0.76 & $ {0.83 \pm 0.04 }$& ${\bf 0.94 \pm 0.06}$\\
        vote& 40 & 0.71 & 0.58 & 0.71 & 0.54 & 0.80 & 0.76 & $ {0.85 \pm 0.04 }$& ${\bf 0.94 \pm 0.06}$\\
        vote& 50 & 0.77 & 0.61 & 0.77 & 0.56 & 0.82 & 0.76 & $ {0.84 \pm 0.04 }$& ${\bf 0.94 \pm 0.05}$\\[0.2cm]
        \midrule
        {\bf \#wins} &  & 4 & 0 & 5 & 0 & 4 & 6 & 8 & {\bf 32} \\[0.1cm]
        {\bf Avg. F1 (All)} &  & 0.68 & 0.66 & 0.68 & 0.64 & 0.76 & 0.63 & 0.81 & \bf{0.90}\\[0.1cm]
        {\bf Avg. F1 (30\%)} &  & 0.66 & 0.64 & 0.66 & 0.64 & 0.74 & 0.58 & 0.80 & \bf{0.89}\\[0.1cm]
        {\bf Avg. F1 (40\%)} &  & 0.69 & 0.65 & 0.68 & 0.65 & 0.76 & 0.65 & 0.80 & \bf{0.90}\\[0.1cm]
        {\bf Avg. F1 (50\%)} &  & 0.70 & 0.68 & 0.70 & 0.65 & 0.79 & 0.66 & 0.83 & \bf{0.90}\\[0.1cm]
        {\bf Avg. Ranking} &  & 4.6 & 5.9 & 4.6 & 5.9 & 3.4 & 5.3 & 2.7 & \bf{1.7}\\[0.1cm]
        \bottomrule
    \end{tabular}
    }
\end{table*}

\paragraph{Statistical tests}
As in the MNIST case, we assess the statistical significance of the results on categorical data using the Friedman statistics and the Nemenyi test \cite{demvsar2006statistical, ienco2016positive}. For the data reported in Table~\ref{tab:categorical f1} the Friedman coefficient is $\chi_{\rm F}^2=146$ with $\mathrm{p-value}=2\cdot 10^{-28}$. Therefore, we reject the null hypothesis of the Friedman test. 

\begin{figure}[!htb]
    \centering
    \includegraphics[width=\columnwidth]{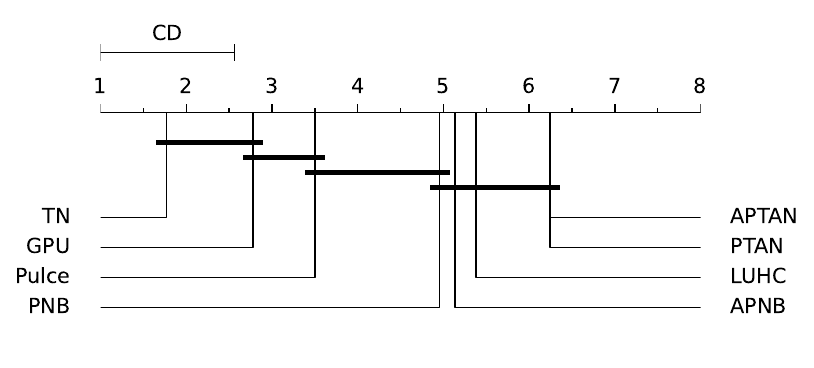}
    \caption{Critical difference graph of the Nemenyi test for the categorical datasets computed using average rankings of methods and 0.05 significance level.}
    \label{fig:nemenyi}
\end{figure}

In the post hoc analysis, we use the Nemenyi test. For $q_{\alpha=0.05}=3.030879$, $N=45$, and $k=8$ the critical distance is $CD=1.56$. Therefore, our method is significantly better than all but the GPU approach (see \fref{fig:nemenyi} for details).

\section{Conclusions and outlook}
\label{sec:conclusion}
We propose a tensor-network approach to the positive unlabeled learning problem and obtain state-of-the-art results on the MNIST image and 15 categorical datasets. To date, no tensor network approach outperformed the best neural network methods on image datasets. However, to use the model on larger images, we need to apply additional preprocessing, e.g., patch-based embedding. Although the proposed loss is complex, all terms in the loss are well-motivated and necessary to obtain good performance and avoid class collapse. We used a co-training-inspired metric (estimated accuracy) for model selection and hyperparameter tuning. We also demonstrated the generation of new positive and negative samples on simple synthetic datasets. Besides missing labels, our model naturally processes also samples with missing attributes. 

We envision two extensions of the proposed approach. First, we can adapt the approach to the general semi-supervised learning problem with any number of classes. Second, based on surprisingly good results with very few (up to one) labeled samples, we expect that we can use the presented approach as an unsupervised clustering method. The idea is to train many one-sample PUL TN classifiers with different labeled samples. We can then compare the trained models with the proposed co-training-like metric to find the number of classes and cluster the dataset.

\section*{Acknowledgement}
The author received support from the Slovenian research agency (ARRS) project J1-2480. Computational resources were provided by SLING – a Slovenian national supercomputing network. The author thanks Dino Ienco for reading the first version of the draft and providing helpful comments.
\medskip
{
\small
\bibliographystyle{unsrt}
\bibliography{PUL_TN}

\begin{thebibliography}{10}

\bibitem{bekker2020learning}
Jessa Bekker and Jesse Davis.
\newblock Learning from positive and unlabeled data: A survey.
\newblock {\em Machine Learning}, 109(4):719--760, 2020.

\bibitem{mordelet2011prodige}
Fantine Mordelet and Jean-Philippe Vert.
\newblock Prodige: Prioritization of disease genes with multitask machine
  learning from positive and unlabeled examples.
\newblock {\em BMC bioinformatics}, 12(1):1--15, 2011.

\bibitem{elkan2008learning}
Charles Elkan and Keith Noto.
\newblock Learning classifiers from only positive and unlabeled data.
\newblock In {\em Proceedings of the 14th ACM SIGKDD international conference
  on Knowledge discovery and data mining}, pages 213--220, 2008.

\bibitem{liu2017computational}
Yashu Liu, Shuang Qiu, Ping Zhang, Pinghua Gong, Fei Wang, Guoliang Xue, and
  Jieping Ye.
\newblock Computational drug discovery with dyadic positive-unlabeled learning.
\newblock In {\em Proceedings of the 2017 SIAM international conference on data
  mining}, pages 45--53. SIAM, 2017.

\bibitem{ren2014positive}
Yafeng Ren, Donghong Ji, and Hongbin Zhang.
\newblock Positive unlabeled learning for deceptive reviews detection.
\newblock In {\em Proceedings of the 2014 conference on empirical methods in
  natural language processing (EMNLP)}, pages 488--498, 2014.

\bibitem{xiao2011similarity}
Yanshan Xiao, Bo~Liu, Jie Yin, Longbing Cao, Chengqi Zhang, and Zhifeng Hao.
\newblock Similarity-based approach for positive and unlabelled learning.
\newblock In {\em Twenty-second international joint conference on artificial
  intelligence}, 2011.

\bibitem{zhou2009learning}
Ke~Zhou, Gui-Rong Xue, Qiang Yang, and Yong Yu.
\newblock Learning with positive and unlabeled examples using topic-sensitive
  plsa.
\newblock {\em IEEE Transactions on Knowledge and Data Engineering},
  22(1):46--58, 2009.

\bibitem{hou2017generative}
Ming Hou, Brahim Chaib-draa, Chao Li, and Qibin Zhao.
\newblock Generative adversarial positive-unlabelled learning.
\newblock In {\em Proceedings of the Twenty-Seventh International Joint
  Conference on Artificial Intelligence, {IJCAI-18}}, pages 2255--2261.
  International Joint Conferences on Artificial Intelligence Organization, 7
  2018.

\bibitem{chiaroni2018learning}
Florent Chiaroni, Mohamed-Cherif Rahal, Nicolas Hueber, and Fr{\'e}d{\'e}ric
  Dufaux.
\newblock Learning with a generative adversarial network from a positive
  unlabeled dataset for image classification.
\newblock In {\em 2018 25th IEEE International Conference on Image Processing
  (ICIP)}, pages 1368--1372. IEEE, 2018.

\bibitem{ienco2016positive}
Dino Ienco and Ruggero~G Pensa.
\newblock Positive and unlabeled learning in categorical data.
\newblock {\em Neurocomputing}, 196:113--124, 2016.

\bibitem{xu2019modeling}
Lei Xu, Maria Skoularidou, Alfredo Cuesta-Infante, and Kalyan Veeramachaneni.
\newblock Modeling tabular data using conditional gan.
\newblock {\em Advances in Neural Information Processing Systems}, 32, 2019.

\bibitem{papivc2023conditional}
Ale{\v{s}} Papi{\v{c}}, Igor Kononenko, and Zoran Bosni{\'c}.
\newblock Conditional generative positive and unlabeled learning.
\newblock {\em Expert Systems with Applications}, page 120046, 2023.

\bibitem{basile2019ensembles}
Teresa Maria~Altomare Basile, Nicola Di~Mauro, Floriana Esposito, Stefano
  Ferilli, and Antonio Vergari.
\newblock Ensembles of density estimators for positive-unlabeled learning.
\newblock {\em Journal of Intelligent Information Systems}, 53(2):199--217,
  2019.

\bibitem{wang2020anomaly}
Jinhui Wang, Chase Roberts, Guifre Vidal, and Stefan Leichenauer.
\newblock Anomaly detection with tensor networks.
\newblock {\em arXiv preprint arXiv:2006.02516}, 2020.

\bibitem{glasser2019expressive}
Ivan Glasser, Ryan Sweke, Nicola Pancotti, Jens Eisert, and Ignacio Cirac.
\newblock Expressive power of tensor-network factorizations for probabilistic
  modeling.
\newblock {\em Advances in neural information processing systems}, 32, 2019.

\bibitem{schollwock2011density}
Ulrich Schollw{\"o}ck.
\newblock The density-matrix renormalization group in the age of matrix product
  states.
\newblock {\em Annals of physics}, 326(1):96--192, 2011.

\bibitem{vzunkovivc2016dynamical}
Bojan {\v{Z}}unkovi{\v{c}}, Alessandro Silva, and Michele Fabrizio.
\newblock Dynamical phase transitions and loschmidt echo in the infinite-range
  xy model.
\newblock {\em Philosophical Transactions of the Royal Society A: Mathematical,
  Physical and Engineering Sciences}, 374(2069):20150160, 2016.

\bibitem{vzunkovivc2018dynamical}
Bojan {\v{Z}}unkovi{\v{c}}, Markus Heyl, Michael Knap, and Alessandro Silva.
\newblock Dynamical quantum phase transitions in spin chains with long-range
  interactions: Merging different concepts of nonequilibrium criticality.
\newblock {\em Physical review letters}, 120(13):130601, 2018.

\bibitem{lerose2018chaotic}
Alessio Lerose, Jamir Marino, Bojan {\v{Z}}unkovi{\v{c}}, Andrea Gambassi, and
  Alessandro Silva.
\newblock Chaotic dynamical ferromagnetic phase induced by nonequilibrium
  quantum fluctuations.
\newblock {\em Physical review letters}, 120(13):130603, 2018.

\bibitem{stoudenmire2016supervised}
Edwin Stoudenmire and David~J Schwab.
\newblock Supervised learning with tensor networks.
\newblock {\em Advances in neural information processing systems}, 29, 2016.

\bibitem{zunkovic2022Deep}
Bojan Žunkovič.
\newblock Deep tensor networks with matrix product operators.
\newblock {\em Quantum Machine Intelligence volume}, 4(21), 2022.

\bibitem{stoudenmire2018learning}
E~Miles Stoudenmire.
\newblock Learning relevant features of data with multi-scale tensor networks.
\newblock {\em Quantum Science and Technology}, 3(3):034003, 2018.

\bibitem{sun2020generative}
Zheng-Zhi Sun, Cheng Peng, Ding Liu, Shi-Ju Ran, and Gang Su.
\newblock Generative tensor network classification model for supervised machine
  learning.
\newblock {\em Physical Review B}, 101(7):075135, 2020.

\bibitem{selvan2021patch}
Raghavendra Selvan, Erik~B Dam, Søren~Alexander Flensborg, and Jens Petersen.
\newblock Patch-based medical image segmentation using matrix product state
  tensor networks.
\newblock {\em Machine Learning for Biomedical Imaging}, 1:1--24, 2022.

\bibitem{vzunkovivc2022grokking}
Bojan {\v{Z}}unkovi{\v{c}} and Enej Ilievski.
\newblock Grokking phase transitions in learning local rules with gradient
  descent.
\newblock {\em arXiv preprint arXiv:2210.15435}, 2022.

\bibitem{yu2002pebl}
Hwanjo Yu, Jiawei Han, and Kevin Chen-Chuan Chang.
\newblock Pebl: positive example based learning for web page classification
  using svm.
\newblock In {\em Proceedings of the eighth ACM SIGKDD international conference
  on Knowledge discovery and data mining}, pages 239--248, 2002.

\bibitem{elkan2001foundations}
Charles Elkan.
\newblock The foundations of cost-sensitive learning.
\newblock In {\em International joint conference on artificial intelligence},
  volume~17, pages 973--978. Lawrence Erlbaum Associates Ltd, 2001.

\bibitem{gan2017bayesian}
Hongxiao Gan, Yang Zhang, and Qun Song.
\newblock Bayesian belief network for positive unlabeled learning with
  uncertainty.
\newblock {\em Pattern Recognition Letters}, 90:28--35, 2017.

\bibitem{chiaroni2019generating}
Florent Chiaroni, Ghazaleh Khodabandelou, Mohamed-Cherif Rahal, Nicolas Hueber,
  and Frederic Dufaux.
\newblock Counter-examples generation from a positive unlabeled image dataset.
\newblock {\em Pattern Recognition}, 107:107527, 2020.

\bibitem{denis2003text}
Francois Denis, Anne Laurent, R{\'e}mi Gilleron, and Marc Tommasi.
\newblock Text classification and co-training from positive and unlabeled
  examples.
\newblock In {\em Proceedings of the ICML 2003 workshop: the continuum from
  labeled to unlabeled data}, pages 80--87, 2003.

\bibitem{zhou2012multi}
Joey~Tianyi Zhou, Sinno~Jialin Pan, Qi~Mao, and Ivor~W Tsang.
\newblock Multi-view positive and unlabeled learning.
\newblock In {\em Asian conference on machine learning}, pages 555--570. PMLR,
  2012.

\bibitem{blum1998combining}
Avrim Blum and Tom Mitchell.
\newblock Combining labeled and unlabeled data with co-training.
\newblock In {\em Proceedings of the eleventh annual conference on
  Computational learning theory}, pages 92--100, 1998.

\bibitem{calvo2007learning}
Borja Calvo, Pedro Larranaga, and Jos{\'e}~A Lozano.
\newblock Learning bayesian classifiers from positive and unlabeled examples.
\newblock {\em Pattern Recognition Letters}, 28(16):2375--2384, 2007.

\bibitem{friedman1997bayesian}
Nir Friedman, Dan Geiger, and Moises Goldszmidt.
\newblock Bayesian network classifiers.
\newblock {\em Machine learning}, 29(2):131--163, 1997.

\bibitem{shao2015laplacian}
Yuan-Hai Shao, Wei-Jie Chen, Li-Ming Liu, and Nai-Yang Deng.
\newblock Laplacian unit-hyperplane learning from positive and unlabeled
  examples.
\newblock {\em Information Sciences}, 314:152--168, 2015.

\bibitem{power2022grokking}
Alethea Power, Yuri Burda, Harri Edwards, Igor Babuschkin, and Vedant Misra.
\newblock Grokking: Generalization beyond overfitting on small algorithmic
  datasets.
\newblock {\em arXiv preprint arXiv:2201.02177}, 2022.

\bibitem{novikov2015tensorizing}
Alexander Novikov, Dmitrii Podoprikhin, Anton Osokin, and Dmitry~P Vetrov.
\newblock Tensorizing neural networks.
\newblock {\em Advances in neural information processing systems}, 28, 2015.

\bibitem{panagakis2021tensor}
Yannis Panagakis, Jean Kossaifi, Grigorios~G Chrysos, James Oldfield, Mihalis~A
  Nicolaou, Anima Anandkumar, and Stefanos Zafeiriou.
\newblock Tensor methods in computer vision and deep learning.
\newblock {\em Proceedings of the IEEE}, 109(5):863--890, 2021.

\bibitem{cichocki2014tensor}
Andrzej Cichocki.
\newblock Tensor networks for big data analytics and large-scale optimization
  problems.
\newblock {\em arXiv preprint arXiv:1407.3124}, 2014.

\bibitem{cichocki2017tensor}
Andrzej Cichocki, Anh-Huy Phan, Qibin Zhao, Namgil Lee, Ivan Oseledets, Masashi
  Sugiyama, Danilo~P Mandic, et~al.
\newblock Tensor networks for dimensionality reduction and large-scale
  optimization: Part 2 applications and future perspectives.
\newblock {\em Foundations and Trends{\textregistered} in Machine Learning},
  9(6):431--673, 2017.

\bibitem{luo2021adjusting}
Xin Luo, Minzhi Chen, Hao Wu, Zhigang Liu, Huaqiang Yuan, and MengChu Zhou.
\newblock Adjusting learning depth in nonnegative latent factorization of
  tensors for accurately modeling temporal patterns in dynamic qos data.
\newblock {\em IEEE Transactions on Automation Science and Engineering},
  18(4):2142--2155, 2021.

\bibitem{chen2022mnl}
Minzhi Chen, Chunlin He, and Xin Luo.
\newblock Mnl: A highly-efficient model for large-scale dynamic weighted
  directed network representation.
\newblock {\em IEEE Transactions on Big Data}, 2022.

\bibitem{zheng2021fully}
Yu-Bang Zheng, Ting-Zhu Huang, Xi-Le Zhao, Qibin Zhao, and Tai-Xiang Jiang.
\newblock Fully-connected tensor network decomposition and its application to
  higher-order tensor completion.
\newblock In {\em Proceedings of the AAAI conference on artificial
  intelligence}, volume~35, pages 11071--11078, 2021.

\bibitem{demvsar2006statistical}
Janez Dem{\v{s}}ar.
\newblock Statistical comparisons of classifiers over multiple data sets.
\newblock {\em The Journal of Machine learning research}, 7:1--30, 2006.

\end{thebibliography}
}

\end{document}